\documentclass[nohyperref]{article}

\usepackage{microtype}
\usepackage{graphicx}
\usepackage{booktabs} 

\usepackage{hyperref}

\usepackage[accepted]{icml2022}

\usepackage{amsmath}
\usepackage{amssymb}
\usepackage{mathtools}
\usepackage{amsthm}

\usepackage{url}
\usepackage{graphicx}
\usepackage{wrapfig}
\usepackage{subcaption}

\usepackage{algorithm}
\usepackage{algorithmic}
\usepackage{multicol}

\usepackage{enumitem}
\usepackage{esvect}

\usepackage[capitalize,noabbrev]{cleveref}

\theoremstyle{plain}
\newtheorem{theorem}{Theorem}[section]
\newtheorem{proposition}[theorem]{Proposition}

\theoremstyle{definition}

\theoremstyle{remark}

\usepackage[textsize=tiny]{todonotes}

\icmltitlerunning{Representation Topology Divergence: a Method for Comparing Neural Network Representations}

\begin{document}

\twocolumn[
\icmltitle{Representation Topology Divergence: a Method for Comparing Neural Network Representations}



\icmlsetsymbol{equal}{*}

\begin{icmlauthorlist}
\icmlauthor{Serguei Barannikov}{skoltech,cnrs}
\icmlauthor{Ilya Trofimov}{skoltech}
\icmlauthor{Nikita Balabin}{skoltech}
\icmlauthor{Evgeny Burnaev}{skoltech,airi}
\end{icmlauthorlist}

\icmlaffiliation{skoltech}{Skolkovo Institute of Science and Technology, Moscow, Russia}
\icmlaffiliation{cnrs}{CNRS, Université Paris Cité, France}
\icmlaffiliation{airi}{Artificial Intelligence Research Institute (AIRI), Moscow, Russia}

\icmlcorrespondingauthor{Serguei Barannikov}{S.Barannikov@skoltech.ru}

\icmlkeywords{Machine Learning, ICML}

\vskip 0.3in
]



\printAffiliationsAndNotice{}  

\begin{abstract}
Comparison of data representations is a complex multi-aspect problem that has no complete solution yet. We propose a method for comparing two data representations. We introduce the Representation Topology Divergence (RTD) which measures the dissimilarity in multi-scale topology between two point clouds of equal size with a one-to-one correspondence between points. The data point clouds are allowed to lie in different ambient spaces. The RTD is one of the few practical methods based on Topological Data Analysis (TDA) applicable to real machine learning datasets. Experiments show the proposed RTD agrees with the intuitive assessment of data representation similarity and is sensitive to its topological structure. We apply RTD to gain insights into neural network representations in computer vision and NLP domains for various problems: training dynamics analysis, data distribution shift, transfer learning, ensemble learning.

\end{abstract}

\section{Introduction}

Representations of objects are the essential component learnt by deep neural networks. In opposite to the distance in the original space, the similarity of representations is proved to be semantically meaningful. Despite the significant practical success of deep neural networks, many aspects of their behavior are poorly understood.
Only a few methods study neural representations without relying on their quality on a specific downstream task.
In this work, we focus on the comparison of representations from neural networks. 

Comparison of representations is an ill-posed problem without a ``ground truth'' answer.
Early studies were based on variants of Canonical Correlation Analysis (CCA): SVCCA, \citep{raghu2017svcca}, PWCCA \citep{morcos2018insights}. However, CCA-like measures define similarity too loosely since they are invariant under any invertible linear transformation. The Centered Kernel Alignment (CKA), \citep{kornblith2019similarity} is the statistical test to measure the independence of two sets of variables. \citep{kornblith2019similarity} proved CKA to be more consistent with the intuitive similarity of representations. Particularly, neural networks learn similar representations from different seeds as evaluated by CKA.
Another line of work is concerned with the alignment between groups of neurons \citep{li2015convergent}, \citep{wang2018towards}. The similarity of representations is also a topic of study in neuroscience \citep{edelman1998representation, kriegeskorte2008representational, connolly2012representation}.

Representational similarity metrics like CKA and CCA were used to gain insights on representations obtained in meta-learning \citep{raghu2019rapid}, to compare representations from different layers of language models \citep{voita2019bottom}, and to study the effect of fine-tuning \citep{wu2020similarity}. Finally, \citep{nguyen2020wide} used CKA to study the phenomenon of a ``block structure'' emerging in wide and deep networks in computer vision and compare their representations. 

In this paper, we take a topological perspective on the comparison of neural network representations.
We propose the \textit{Representation Topology Divergence (RTD)} score, which measures dissimilarity between two point clouds of equal size
with a one-to-one correspondence between points. Point clouds are allowed to lie in different ambient spaces. 
Existing geometrical and topological methods are dedicated to other problems: they are either too general and do not incorporate the one-to-one correspondence requirement  \citep{khrulkov2018geometry}, \citep{tsitsulin2019shape}, or they restrict point clouds to lie in the same ambient space \citep{kynkaanniemi2019improved}, \citep{barannikov2021manifold}. 
Most of these methods 
are applied to the evaluation of GANs. 
Recently, \citep{moor2020topological} proposed a loss term to compare the topology of data in original and latent spaces 
and applied the term as a part of the Topological Autoencoder.




\begin{figure*}[tp]
\centering
\begin{subfigure}{.49\linewidth}
  \centering
  \includegraphics[width=1\linewidth]{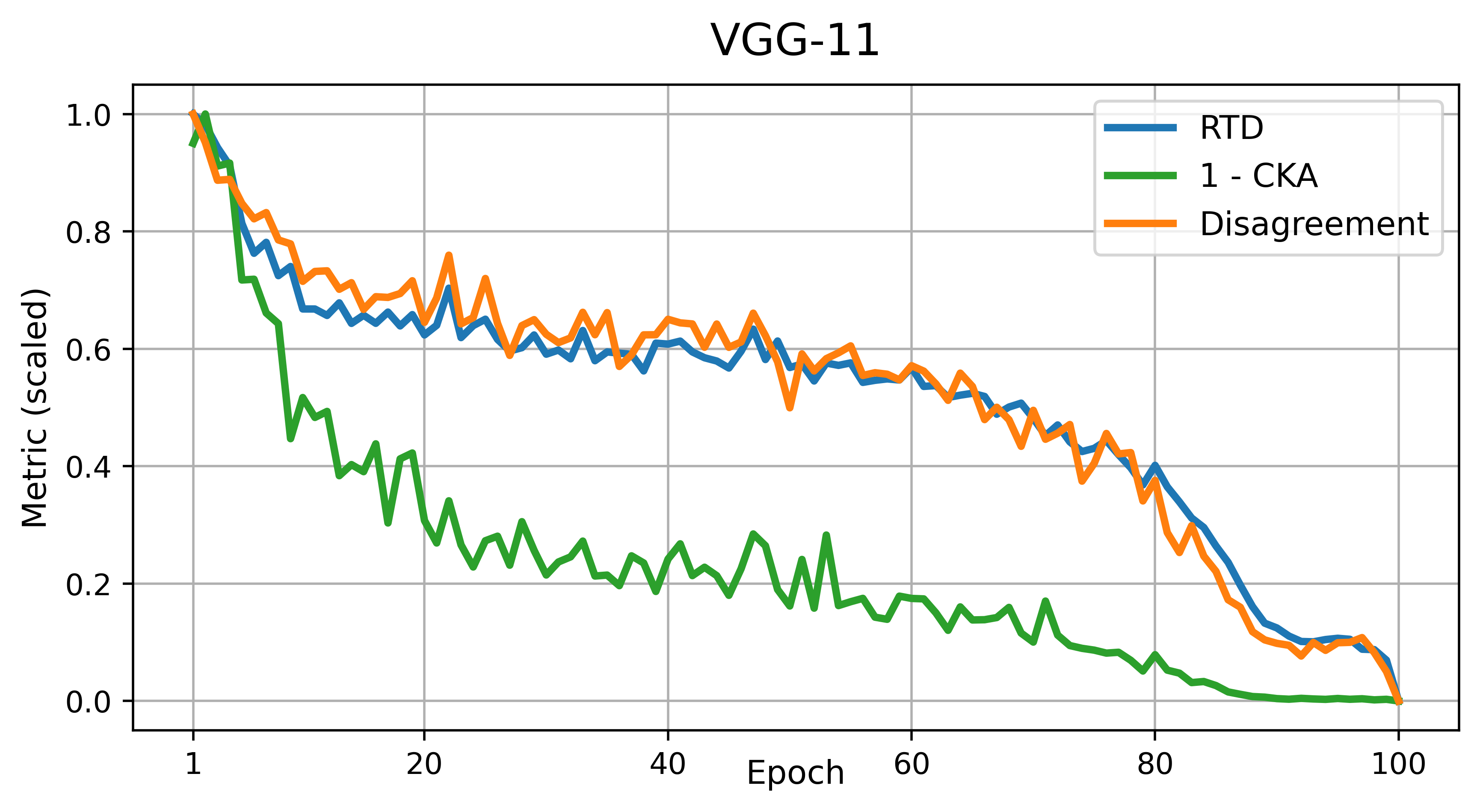}
\end{subfigure}%
\hfill
\begin{subfigure}{.49\linewidth}
  \centering
  \includegraphics[width=1\linewidth]{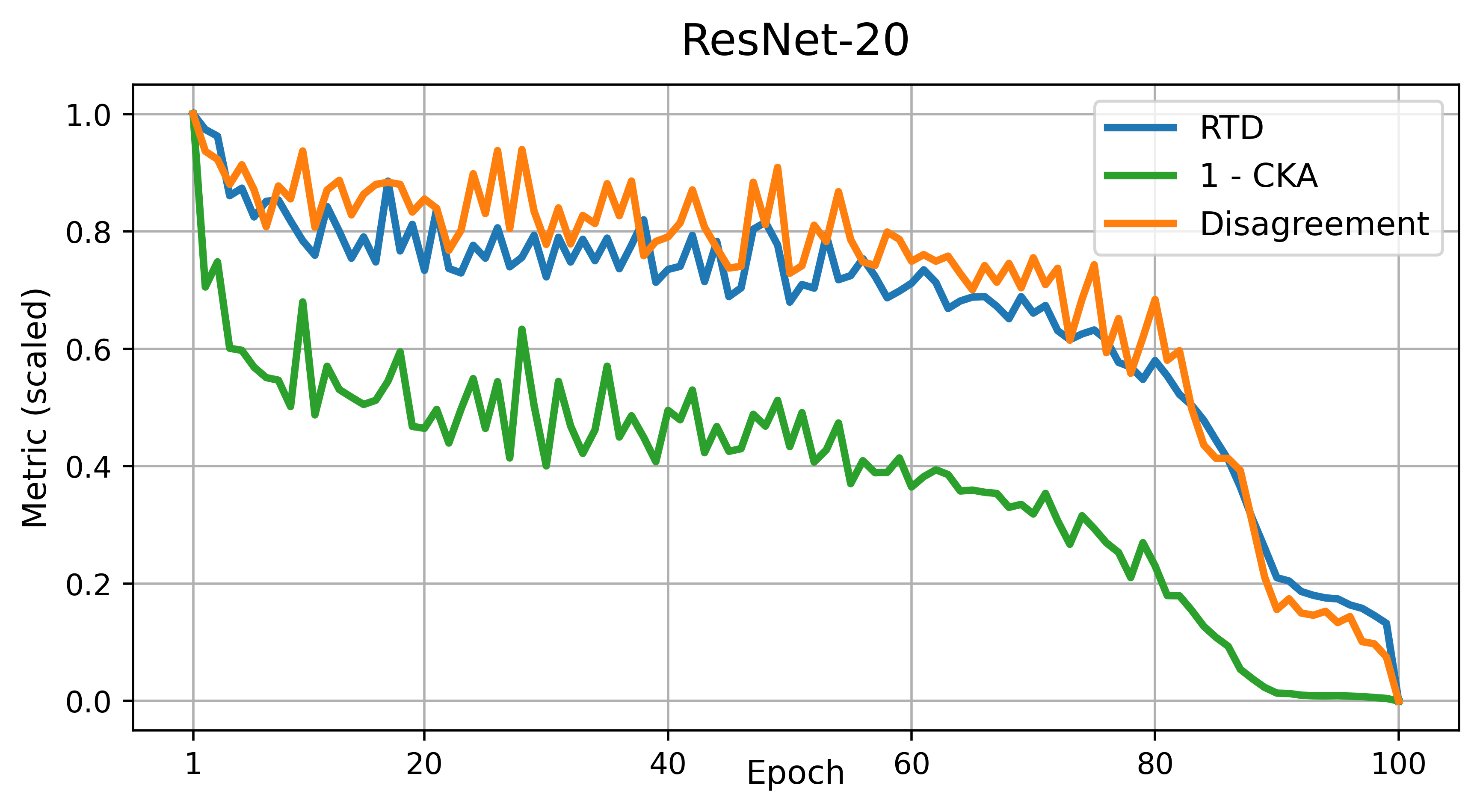}
\end{subfigure}
\caption{
Comparison of representations after the $i$th epoch and the final one done by RTD, $1-$CKA, and disagreement of predictions. All the measures are normalized by division to their maximal values. Strikingly, RTD highly correlates with the disagreement of models' predictions.
}
\label{fig:evolution}
\end{figure*}

In this work, we make the following contributions:
\begin{enumerate}
    \item We propose a topologically-inspired approach for comparison of neural network representations;
    \item We introduce the \textit{R-Cross-Barcode$(P, \tilde{P})$}, a tool based on Topological Data Analysis (TDA), which measures the differences in the multi-scale topology of two point clouds $P, \tilde{P}$ with a one-to-one correspondence between points;
    \item Based on the \textit{R-Cross-Barcode$(P, \tilde{P})$}, we define the \textit{Representation Topology Divergence (RTD)}, the quantity measuring the multi-scale topological dissimilarity between two representations;
    \item Our computational experiments show that RTD agrees with an intuitive notion of neural network representations similarity. In contrast to most existing approaches, RTD is sensitive to differences in topological structures (clusters, voids, cavities, tunnels, etc.)  of the representations and enjoys a very good correlation with disagreement of models predictions. We apply RTD to compare representations in computer vision and NLP domains and various problems: training dynamics analysis, data distribution shift, transfer learning, ensemble learning, and disentanglement. Experiments show that RTD outperforms CKA, IMD, and SVCCA.
\end{enumerate}

The source code is publicly available:\\ \url{https://github.com/IlyaTrofimov/RTD}.

\section{Comparing Neural Network Representations}

Our starting point is the geometric perspective on representation learning through the lens of the manifold hypothesis \citep{goodfellow2016deep}, according to which real-world data
presented in a high-dimensional space are expected to concentrate in the vicinity of a manifold of much lower
dimension. The low-dimensional manifold $M_{\cal{P}}$ underlying the given data representation ${\cal{P}}$ can be accessed in general only through discrete sets of samples. The standard approach to recover the manifold $M_{\cal{P}}$ is to take a sample $P$ and to approximate $M_{\cal{P}}$ by a set of simplexes with vertices from $P$. Commonly, to select the simplexes approximating $M_{\cal{P}}$ one has to fix a threshold $\alpha>0$ and consider the simplexes with edge lengths not exceeding $\alpha$ \citep{niyogi2008finding,belkin2001laplacian}. It is difficult to guess the correct value of the threshold, and hence a reasonable approach is to study all thresholds at once.


Given two representations, we consider two corresponding graphs with distance-like weights and compare the difference in the multiscale topology of the two graphs.

Let $\cal{P},\cal{\tilde{P}}$ be two representations giving two embeddings of the same data $\cal{V}$. The two embeddings ${\cal{P}},\tilde{{\cal{P}}}$ belong in general to different ambient spaces and have the natural one-to-one correspondence between points in ${\cal{P}}$ and $\tilde{{\cal{P}}}$. 
Given a sample of data $V\subseteq \cal{V}$, 
the two representations $P={\cal{P}}(V)$, $\tilde{P}=\tilde{\cal{P}}(V)$ define two weighted graphs $\mathcal{G}^{{w}}$, $\mathcal{G}^{\tilde{w}}$ with the same vertex set $V$.  The weights $w_{AB}$, $\tilde{w}_{AB}$  of an edge $AB$ are given by the distances $w_{AB}={\operatorname{dist}}(P(A),P(B))$, $\tilde{w}_{AB}={\operatorname{dist}} (\tilde{P}(A),\tilde{P}(B))$. 

The simplicial approximation to the manifold $M_{\cal{P}}$ at threshold $\alpha$ consists of simplexes whose edges in $\mathcal{G}^{w}$ have weights not exceeding $\alpha$. Let $\mathcal{G}^{{w}\leq \alpha}$ denote the graph with the vertex set $V$ and the edges with weights not exceeding $\alpha$. 
To compare the simplicial approximations to the manifolds $M_{\cal{P}}$ and  $M_{\tilde{\cal{P}}}$  described by the graphs $\mathcal{G}^{{w}\leq \alpha}$ and $\mathcal{G}^{\tilde{w}\leq \alpha}$, we compare each of the two simplicial approximations with the union of simplices formed by edges present in at least one of the two graphs.  
The graph $\mathcal{G}^{\min(w,\tilde{w})\leq \alpha}$ contains an edge between vertices 
$A$ and $B$ iff the distance between the points $A$ and $B$ is smaller than $\alpha$ in at least one of the representations $P$, $\tilde{P}$. 
The set of edges of the graph $\mathcal{G}^{\min(w,\tilde{w})\leq \alpha}$ is the union of sets of edges of $\mathcal{G}^{{w}\leq \alpha}$ and $\mathcal{G}^{\tilde{w}\leq \alpha}$.  The similarity of manifolds  
$M_{\cal{P}}$ and  $M_{\tilde{\cal{P}}}$ can be measured by the degrees of similarities of the graph  $\mathcal{G}^{\min(w,\tilde{w})\leq \alpha}$ with the graph $\mathcal{G}^{{w}\leq \alpha}$ and the graph $\mathcal{G}^{\tilde{w}\leq \alpha}$. 
 \begin{figure}[h!]
    \centering
    \vskip-0.05in
    \includegraphics[width=0.46\textwidth]{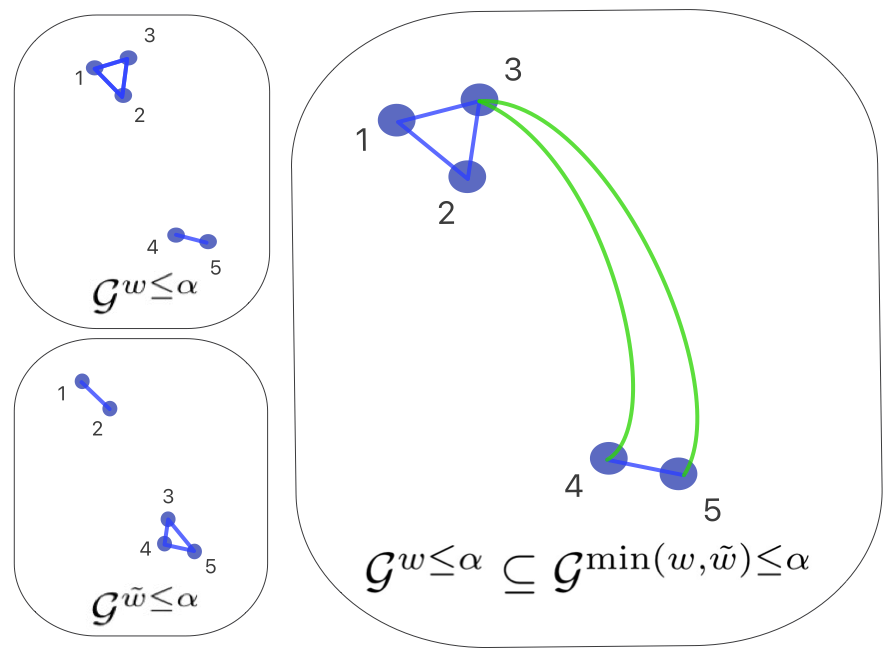}
     \caption{Graphs $\mathcal{G}^{{w}\leq \alpha}$, $\mathcal{G}^{\tilde{w}\leq \alpha}$ and  $\mathcal{G}^{\min(w,\tilde{w})\leq \alpha}$ with  edges not in $\mathcal{G}^{{w}\leq \alpha}$ colored in green.}\label{fig:graphmin}
    \label{fig:2graphs}
    \vskip-0.2in
\end{figure}

\subsection{Topological features for a pair of weighted graphs}\label{sec:Topfeatures}
One way to measure the discrepancy between the graphs $\mathcal{G}^{{w}\leq \alpha}$ and  $\mathcal{G}^{\min(w,\tilde{w})\leq \alpha}$ is to count the graph $\mathcal{G}^{{w}\leq \alpha}$ connected components merged together in the graph $\mathcal{G}^{\min(w,\tilde{w})\leq \alpha}$. We show an example of this situation in Figure \ref{fig:graphmin} right, see also Figure \ref{fig:3to2clusters}, where three graphs $\mathcal{G}^{{w}\leq \alpha}$, $\mathcal{G}^{\tilde{w}\leq \alpha}$ and  $\mathcal{G}^{\min(w,\tilde{w})\leq \alpha}$ are shown, with edges of the graph $\mathcal{G}^{\min(w,\tilde{w})\leq \alpha}$ not in $\mathcal{G}^{{w}\leq \alpha}$ colored in green.  Each merging is represented by a class of green paths in $\mathcal{G}^{\min(w,\tilde{w})\leq \alpha}$  joining two  blue clusters. The significance of the discrepancy constituted by the green path is measured by the  difference $\alpha_d-\alpha_b$ in the smallest thresholds $\alpha_b,\alpha_d$ at which the two clusters are  merged in $\mathcal{G}^{\min(w,\tilde{w})\leq \alpha_b}$ and $\mathcal{G}^{{w}\leq \alpha_d}$.
Homology is the tool that permits counting such topological features, because of the space limit we gather the definitions and necessary properties of homology in Appendix \ref{app:background}, see also \citep{hatcher2005algebraic}. 
The number of these simplest topological features is the dimension of the kernel of linear map $H_0(\mathcal{G}^{{w}\leq \alpha})\to H_0(\mathcal{G}^{\min(w,\tilde{w})\leq \alpha})$, as basis elements of the vector space $H_0$ correspond to the graph connected components. It may also happen that a non-trivial merging happens between two distant parts of the same $\mathcal{G}^{{w}\leq \alpha}$ cluster or between two $\mathcal{G}^{{w}\leq \alpha}$ clusters already connected via a chain of merging, as on Figure \ref{fig:5clusters}. The number of these features is the dimension of the cokernel of the map $H_1(\mathcal{G}^{{w}\leq \alpha})\to H_1(\mathcal{G}^{\min(w,\tilde{w})\leq \alpha})$. Hence the number of non-trivial mergings is the sum of the two numbers.
We are interested in these numbers for all possible thresholds $\alpha$.
When the threshold $\alpha$ is increased then more green and blue edges appear, and also certain green edges become blue. Using an auxiliary graph and the barcodes algorithm, we calculate the numbers of such topological features for all values of $\alpha$ at once.

\begin{figure*}[t!]
\begin{multicols}{2} \setlength{\tabcolsep}{1pt}
\begin{algorithm}[H]  
   \caption{$\text{R-Cross-Barcode}_{i}(P,\tilde{P})$}\label{alg:RcrossB} 
\begin{algorithmic}
   \STATE\hskip-1em {\bfseries Input:} $w$, $\tilde{w}$ : matrices of pairwise distances within point clouds $P$, $\tilde{P}$  
   \STATE\hskip-1em {\bfseries Require:} $\text{vr}(m)$: function computing  filtered complex from pairwise distances matrix \nolinebreak$m$
   \STATE\hskip-1em {\bfseries Require:} $\text{B}(C,i)$: function computing persistence intervals of filtered complex $C$ in dimension $i$
   \STATE $w$, $\tilde{w}\gets$ $w$, $\tilde{w}$ divided by their 0.9 quantiles
   \STATE 
    $m\gets \begin{pmatrix}
      w & (w_+)^\intercal & 0 \\\
      w_+ & \min(w,\tilde{w}) & +\infty\\\ 0 & +\infty & 0
         \end{pmatrix}$
   \STATE \emph{R-Cross-Barcode}$_{i}\gets \text{B}(\text{vr}(m),i)$
   \STATE \hskip-1em{\bfseries Return:} intervals list \emph{R-Cross-Barcode}$_{i} (P,\tilde{P})$ representing "births" and "deaths" of topological discrepancies between $P$  and $\tilde{P}$.
\end{algorithmic}
\end{algorithm}
\columnbreak
\begin{algorithm}[H]
  \caption{\emph{RTD}($\cal{P},\cal{\tilde{P}}$), see section \ref{sec:alg} for details, suggested default values: $b=500$, $n=10$ }
   \label{alg:Rtop}
\begin{algorithmic}
   \STATE \hskip-1em {\bfseries Input:} ${\cal{P}}\in {\mathbb{R}}^{\lvert \mathcal{V}\rvert\times D}$, ${\cal{\tilde{P}}}\in {\mathbb{R}}^{\lvert \mathcal{V}\rvert\times \tilde{D}}$ : data representations
   \FOR{$j=1$ {\bfseries to} $n$}
     \STATE $V_j\gets$ random choice ($\mathcal{V},b$)
     \STATE $P_j,\tilde{P}_j\gets$ ${\cal{P}}(V_j),{\cal{\tilde{P}}}(V_j)$
     \STATE ${\cal{B}}_j\gets$ \emph{R-Cross-Barcode}$_{1}(P_j,\tilde{P}_j)$ intervals' list calculated by Algorithm \ref{alg:RcrossB}
     \STATE $rtd_j\gets$ sum of lengths of all intervals in ${\cal{B}}_j$
   \ENDFOR
   \STATE \emph{RTD}$_1({\cal{P}},{\cal{\tilde{P}}})\gets\text{mean}(rtd)$
   \STATE \hskip-1em{\bfseries Return:} number \textbf{\emph{RTD}}$_1(\cal{P},\cal{\tilde{P}}$)
   representing discrepancy between the representations $\cal{P},\cal{\tilde{P}}$
\end{algorithmic}
\end{algorithm}
\end{multicols}
\vskip-2em
\end{figure*}

\subsection{R-Cross-Barcode}\label{subs:RCrB}
Recall that the Vietoris-Rips filtered complex of a graph $\mathcal{G}$ equipped with edge weights' matrix $m$ is the collection of  $k-$simplexes, $k\geq 0$, which are $(k+1)-$element subsets of the set of vertices of $\mathcal{G}$, with the filtration threshold of a simplex defined by the maximal weight on the edges:   
\begin{equation*}
   R_\alpha({\mathcal{G}}^m)=\left\{\{A_{i_0},\ldots,A_{i_k}\}, A_i\in\operatorname{Vert}({\mathcal{G}})\vert m_{A_{i}A_{j}}\leq \alpha \right\}
\end{equation*}
Our simplicial approximation to the manifold $M_{\mathcal{P}}$ at threshold $\alpha$ is the union of all simplexes from the simplicial complex $R_\alpha(\mathcal{G}^w)$, and similarly the approximation to $M_{\tilde{\cal{P}}}$ is the union of all simplexes from $R_\alpha(\mathcal{G}^{\tilde{w}})$. 

The dissimilarity between the filtered simplicial complexes $R_\alpha(\mathcal{G}^w)$ and $R_\alpha(\mathcal{G}^{\tilde{w}})$ can be quantified using 
the homological methods. The relevant tools here are  homology,  barcodes and homology exact sequences. We describe our construction below and, because of space limitations, we sketch further explanation of the construction in Appendix, Section \nolinebreak \ref{propRcross}. 




Concretely, to compare the multi-scale topology of the two weighted graphs $\mathcal{G}^{w}$ and  $\mathcal{G}^{\tilde{w}}$ we introduce the
weighted graph 
$\hat{\mathcal{G}}^{w,\tilde{w}}$ with doubled set of vertices and with the edge weights defined as follows. For convenience, fix a numbering of vertices $\text{Vert}(\mathcal{G})=\{A_1,\ldots,A_N\}$.
For each vertex $A\in \text{Vert}(\mathcal{G})$ we add the extra vertex $A'$ together with $A$ 
to $\hat{\mathcal{G}}$, plus the unique additional vertex $O$, and define the distance-like edge weights in $\hat{\mathcal{G}}^{w,\tilde{w}}$ as:
\begin{gather}
    d_{A'_iA'_j}=\min(w_{A_iA_j},\tilde{w}_{A_iA_j}), \,\, d_{A_iA'_j}=d_{A_iA_j}=w_{A_iA_j} , \nonumber\\  d_{A_iA'_i}=d_{OA_i}=0,\,\, d_{A_jA'_i}=d_{OA'_i}=+\infty \label{eq:matrO}
\end{gather}
where $i<j$ and $O\in \text{Vert}(\hat{\mathcal{G}}^{w,\tilde{w}})$ is the additional vertex. In practice, for the calculation of RTD described below, the distance matrix can be taken in a slightly simpler form 
$
m= \begin{pmatrix}
  0 & (w_+)^\intercal \\\
 w_+ & \min(w,\tilde{w}) 
 \end{pmatrix}
$,
where $w$ and $\tilde{w}$ are the edge weight matrices of $\mathcal{G}^w$
and $\mathcal{G}^{\tilde{w}}$, and $w_+$, respectively $(w_+)^\intercal$, is the matrix $w$ with upper-(respectively, lower-\nolinebreak)\nolinebreak triangular part replaced by $+\infty$.


Next, we construct the Vietoris-Rips filtered simplicial complex of the graph $\hat{\mathcal{G}}^{w,\tilde{w}}$ and take its barcode. The doubling of vertices in $ \hat{\mathcal{G}}^{w,\tilde{w}}$ creates triangles $OA_iA_j$, $A_iA_jA'_j$, $A_iA'_iA'_j$ at the threshold $\alpha=w_{A_iA_j}$. These triangles "kill" the edge $A'_iA'_j$ becoming blue at this threshold. Intuitively, the $i-$th barcode of $R_\alpha(\hat{\mathcal{G}}^{w,\tilde{w}})$  records the 
 $i$-dimensional topological features that are born in $\mathcal{G}^{\min(w,\tilde{w})\leq \alpha}$ but are not yet born near the same place in $\mathcal{G}^{{w}\leq\alpha}$  and the $(i-1)-$dimensional topological features that are dead in $\mathcal{G}^{\min(w,\tilde{w})\leq \alpha}$ but are not yet dead at the same place in $\mathcal{G}^{{w}\leq\alpha}$, see Theorem \ref{thr:basic} below.



\textbf{Definition}. The \emph{R-Cross-Barcode$_i(P,\tilde{P})$} is the set of intervals recording  the  “births"  and “deaths"   of $i$-dimensional   topological   features in the filtered simplicial complex $R_\alpha(\hat{\mathcal{G}}^{w,\tilde{w}})$.

The \emph{R-Cross-Barcode$_*(P,\tilde{P})$} (for   \emph{Representations' Cross-Barcode})
records the differences in the multiscale topology of the two embeddings. The topological features with longer lifespans indicate in general the essential features.

\begin{theorem}\label{thr:basic} Basic properties of {R-Cross-Barcode$_*(P,\tilde{P})$}: 
\begin{itemize}[topsep=0pt,noitemsep,nolistsep, partopsep=0pt, parsep=0ex, leftmargin=*]
\item if $P(A)=\tilde{P}(A)$ for any object $A\in V$, then {R-Cross-Barcode}$_*(P,\tilde{P})=\varnothing$;
\item if all distances within $\tilde{P}(V)$ are zero i.e. all objects are represented  by the same point in $\tilde{P}$, then for all $k\ge0$: {R-Cross-Barcode$_{k+1}(P,\tilde{P})=\text{Barcode}_{k}(P)$} the standard barcode of the point cloud $P$;

    \item for any value of threshold $\alpha$, the following sequence of natural linear maps of homology groups
\begin{multline}
    \xrightarrow{r_{3i+3}} H_{i}(R_\alpha(\mathcal{G}^w)) \xrightarrow{r_{3i+2}} H_i(R_\alpha(\mathcal{G}^{\min(w,\tilde{w})}))\xrightarrow{r_{3i+1}}\\ \xrightarrow{r_{3i+1}} H_i(R_\alpha(\hat{\mathcal{G}}^{w,\tilde{w}})) \xrightarrow{r_{3i}} H_{i-1}(R_\alpha(\mathcal{G}^w))\xrightarrow{r_{3i-1}}\\\xrightarrow{r_{3i-1}}\ldots\xrightarrow{r_1} H_{0}(R_\alpha(\mathcal{G}^{\min(w,\tilde{w})}))\xrightarrow{r_0}0\label{eq:longseq}
\end{multline} is exact, i.e. for any $j$ the kernel of the map $r_{j}$ is the image of the map $r_{j+1}$.

\end{itemize}
\end{theorem}
The proof of the first two properties is immediate and the third property follows from the properties of distinguished triangles of complexes, see Appendix \ref{app:background} for more details. The exactness of the sequence (\ref{eq:longseq}) for  $j=1,2,3$ implies that the calculation of the topological features from Section \ref{sec:Topfeatures} for all $\alpha$ is reduced to the calculation of
$H_1(R_\alpha(\hat{\mathcal{G}}^{w,\tilde{w}}))$ for all $\alpha$, i.e. to the calculation of  \emph{R-Cross-Barcode$_1(P,\tilde{P})$}.

\subsection{Representation Topology Divergence.}
The \emph{R-Cross-Barcode$_*(P,\tilde{P})$} is by itself, to our opinion,  a precise and intuitive tool for understanding discrepancies between two representations.  There are several numerical characteristics measuring the non-emptyness of  \emph{R-Cross-Barcode}. Based on experiments and on relation of sum of bars' lengths with Earth Moving Distance \citep{barannikov2021manifold}, we define the sum of lengths of the bars in \emph{R-Cross-Barcode$_i(P,\tilde{P})$}, denoted {$RTD_i(P,\tilde{P})$}, as the scalar characterizing the degree of topological discrepancy between the representations  $P,\tilde{P}$. We use most often the average of  $RTD_1(P,\tilde{P})$ and $RTD_1(\tilde{P},P)$, denoted \emph{RTD} score, in our computations below.

\begin{proposition}
If  $RTD_i(P,\tilde{P})=RTD_i(\tilde{P},P)=0$ for all $i\geq 1$, then the barcodes of the weighted graphs $\mathcal{G}^w$ and $\mathcal{G}^{\tilde{w}}$ are the same in any degree. Moreover, in this case the topological features are located in the same places: 
the inclusions $R_\alpha(\mathcal{G}^w)\subseteq R_\alpha(\mathcal{G}^{\min(w,\tilde{w})})$,  $R_\alpha(\mathcal{G}^{\tilde{w}}) \subseteq R_\alpha(\mathcal{G}^{\min(w,\tilde{w})})$ induce homology isomorphisms for any threshold~$\alpha$.
\end{proposition}

\subsection{Algorithm} \label{sec:alg}
\begin{figure*}[th]
\centering
\begin{subfigure}[t]{\textwidth}
\includegraphics[width=\textwidth]{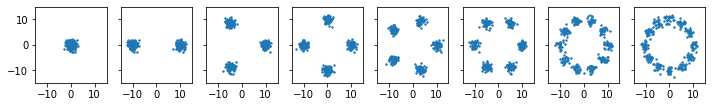}
\caption{Point clouds used in ``clusters'' experiment.}
\end{subfigure}
\label{fig:clusters}

\begin{subfigure}[t]{\textwidth}
\centering
\fcolorbox{red}{white}{
\includegraphics[width=0.23\textwidth]{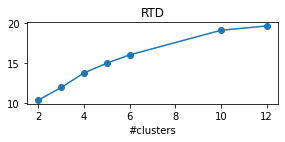}
}
\includegraphics[width=0.23\textwidth]{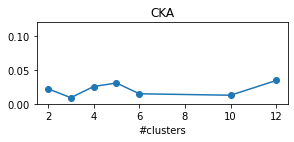}
\includegraphics[width=0.23\textwidth]{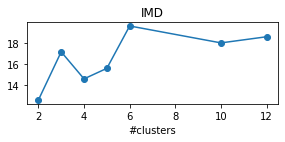}
\includegraphics[width=0.23\textwidth]{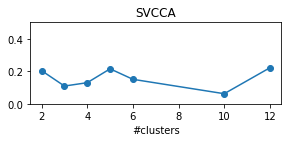}

\caption{Representations' comparison measures. Ideally, the measure should  change monotonically with the increase of topological discrepancy.}
\label{fig:clusters_metrics}
\end{subfigure}

\begin{subfigure}[t]{\textwidth}
\includegraphics[width=1\textwidth]{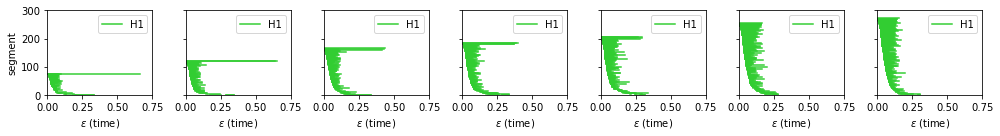}
\caption{R-Cross-Barcode$(P,\tilde{P})$ for the ``clusters'' experiments. $\tilde{P}$ - is the point cloud having one cluster, ${P}$ - 2, 3, 4, 5, 6, 10, 12 clusters.}
\label{fig:clusters_barcodes}
\end{subfigure}

\caption{RTD perfectly detects cluster structures, while rival measures fail. One cluster is compared with 2-12 clusters.}

\label{fig:clusters2}
\end{figure*}

First we compute the \emph{R-Cross-Barcode$_1(P,\tilde{P})$} on two representations $P,\tilde{P}$ of a sample $V$.  For this we calculate the matrices of pairwise distances $w$, $\tilde{w}$ within the point clouds $P$, $\tilde{P}$. We assume that the metrics in the ambient spaces of representations are normalized so that the two point clouds are of comparable size, namely their $0.9$ quantile of pairwise distances coincide. This ensures that our score has scaling invariance, the reasonable property of a good representation similarity measure, as argued in e.g. \citep{kornblith2019similarity}. 
Next, the algorithm builds the Vietoris-Rips complex from the matrix $m$ defined in Equation \ref{eq:matrO}. Then the $1-$dimensional barcode, see \citep{barannikov2021canonical}, of the built filtered simplicial complex is calculated. The last two steps can be done using scripts that are optimized for GPU acceleration \citep{zhang2020gpu}. Then we sum the lengths of bars in  \emph{R-Cross-Barcode$_1(P,\tilde{P})$}. To get the symmetric measure we usually take the half-sum with the similar sum of bars in \emph{R-Cross-Barcode$_1(\tilde{P},P)$}. The computation is repeated a~sufficient number of times to obtain the mean of the chosen characteristics. We have observed experimentally that about $10$ times is usually sufficient for common datasets. 
The main steps of the computation are summarized in Algorithms \ref{alg:RcrossB} and \ref{alg:Rtop}.


%
%

\textbf{Complexity.} Algorithm 1 starts with computation of the two matrices of pairwise distances $w$, $\tilde{w}$ for a pair of representations of a sample $V$: $P\in {\mathbb{R}}^{b\times D}$, $\tilde{P}\in {\mathbb{R}}^{b\times \tilde{D}}$ involving $O(|V|^2(D+\tilde{D}))$ operations. Next, persistent intervals of the filtered complex must be computed. Given the distance matrix $m$, the complexity of their computation does not depend on the dimensions $D,\tilde{D}$ of the data representations.
Generally, the barcode computation is at worst cubic in the number of simplexes involved. In practice, the computation is quite fast since the boundary matrix is typically sparse for real datasets. For R-Cross-Barcodes' calculation, we used  GPU-optimized software. 
Thus, the computation of R-Cross-Barcode takes a similar time as in the previous step even on datasets of high dimensionality.
Since only the dissimilarities in representation topology are calculated, the results are quite robust and a rather low number of iterations is needed to obtain accurate results.

\section{Experiments}

In the experimental section, we study the ability of the proposed R-Cross-Barcodes and RTD to detect changes in topological structures with the use of synthetic point clouds; we demonstrate the superiority of RTD over CKA, SVCCA, IMD (Section \ref{sec:synthetic}). RTD meaningfully compares representations from UMAP with different parameters (Section \ref{sec:umap_mnist}).
By comparing representations from various architectures (Section \ref{sec:nas-bench-nlp}), layers, epochs, ensembles and after data distribution shift (Section \ref{sec:cnn}) we show that RTD is in line with natural notion of representational similarity. A high correlation between RTD and disagreement of neural network predictions is an interesting empirical finding. 
\vspace{-2mm}

\subsection{Experiments with synthetic point clouds}
\label{sec:synthetic}
\vspace{-0.5mm}
We start with small-scale experiments with synthetic point clouds: ``clusters'' and ``rings''. 
For the \textbf{``clusters'' experiment} (Figure \ref{fig:clusters2}, top), the initial point cloud consists of 300 points randomly sampled from the 2-dimensional normal distribution having mean $(0,0)$. Next, we split it into 2,3\ldots12 parts (clusters) and move them to the circle of radius 10. Then, we compare the initial point cloud (having one cluster) with the split ones.

We compared these point clouds  by calculating:
RTD, CKA \citep{kornblith2019similarity}, IMD \citep{tsitsulin2019shape} and SVCCA \citep{raghu2017svcca}. We calculated linear CKA since \citep{kornblith2019similarity} concluded that it provides the same performance as the RBF kernel, but does not require selecting a kernel width. For SVCCA, we calculated average correlation $\bar{\rho}$ for the truncation threshold 0.99, as recommended in  \citep{raghu2017svcca}. The IMD score \citep{tsitsulin2019shape} was very noisy and we averaged it over 100 runs.

\begin{figure*}[t]
\centering
\begin{subfigure}{0.55\textwidth}
\includegraphics[width=0.32\textwidth]{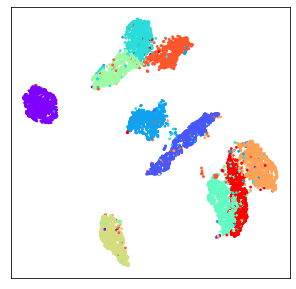}
\includegraphics[width=0.32\textwidth]{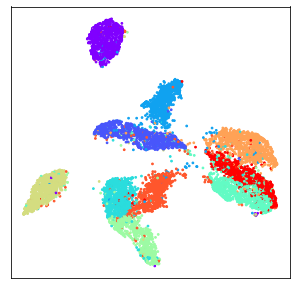}
\includegraphics[width=0.32\textwidth]{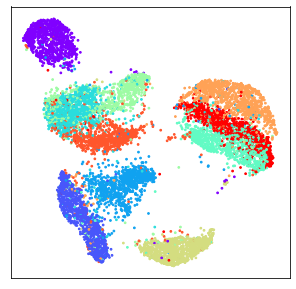}
\caption{2D representations of MNIST with n\_neighbors = 10, 50, 200}
\label{fig:umap_mnist_3vars}
\end{subfigure}
\begin{subfigure}{0.19\textwidth}
\includegraphics[width=\textwidth]{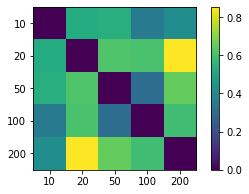}
\caption{1-CKA}
\label{fig:umap_mnist_cka}
\end{subfigure}
\begin{subfigure}{0.19\textwidth}
\includegraphics[width=\textwidth]{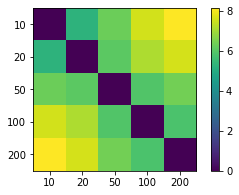}
\caption{RTD}
\end{subfigure}
\caption{Comparing representations of MNIST by UMAP with varying n\_neighbors.}
\label{fig:umap_mnist_cka_rtd}
\end{figure*}

Figure \ref{fig:clusters_metrics} presents the results: RTD perfectly tracks the change of the topological complexity while the alternative measures mostly fail. The Kendall-tau rank correlations of the measures with a number of clusters are: RTD: 1.0, CKA: 0.23, IMD: 0.43, SVCCA: 0.14. We also note that RTD does not have any tunable parameters as SVCCA and does not require averaging over as many runs as IMD.
Figure \ref{fig:clusters_barcodes} shows the $H_1$ R-Cross-Barcodes calculated while comparing clusters. In accordance with the definition of RTD, $H_0$ barcodes are absent. The sum of the lengths of the segments increases with increasing differences in topology. 
Running times and all of the R-Cross-Barcodes are shown in Appendix \ref{app:synthetic-r-cross-barcodes}. Additional representation similarity measures were evaluated in Appendix \ref{app:additional}.

In the \textbf{``rings'' experiment}, we compared synthetic point clouds consisting of a variable number of rings, see Figure \ref{fig:rings_clouds} in Appendix \ref{app:rings}. Initially, there are 500 points uniformly distributed over the unit circle. Then, the points are moved onto circles with radii varying from $0.5$ to $1.5$. Finally, we compare the point cloud having 5 rings with other ones. Figure \ref{fig:rings_metrics} in Appendix \ref{app:rings} present the results. RTD almost ideally reflects the change of the topological complexity while the alternative measures mostly fail. The Kendall-tau rank correlations of the measures with a number of rings are: RTD: 0.8, CKA: -0.2, IMD: 0.8, SVCCA: -0.2.

In the next sections, we compare RTD only with CKA, since it is the most popular method for comparing neural representations \citep{kornblith2019similarity, nguyen2020wide}.

\subsection{Comparing representations from UMAP}
\label{sec:umap_mnist}

UMAP \cite{mcinnes2018umap} is the state-of-the-art method for visualizing high-dimensional datasets by obtaining their 2D/3D representations. We apply UMAP to the MNIST dataset to get 2D representations. We vary the number of neighbors in UMAP in the range $(10, 20, 50, 100, 200)$, see Figure \ref{fig:umap_mnist_3vars} (all of the figures are in Appendix \ref{app:mnist_umap}). This parameter affects the cluster structure: for low values, the algorithm focuses on the local structure and clusters are crisp; for high values, the algorithm pays more attention to the global structure, and clusters were found to often overlap. Then, we perform the pairwise comparison of all the variants of 2D representations by RTD and CKA, see Figure \ref{fig:umap_mnist_cka_rtd}. RTD reveals a nice monotonic pattern w.r.t. a number of neighbors, while values of CKA are quite chaotic. 

\subsection{Experiments with NAS-Bench-NLP}
\label{sec:nas-bench-nlp}

\begin{figure}[h]
    \centering
    \vskip-0.2in
    \includegraphics[width=0.3\textwidth]{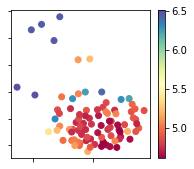}
    \caption{Multi-dimensional scaling of 90  architectures selected randomly from NAS-Bench-NLP. Color depicts log. perplexity.}
    \label{fig:nas-bench-nlp}
    \vskip-0.1in
\end{figure}

Recently, neural architecture search has attracted a lot of attention in the machine learning community 
\citep{liu2019darts, dong2019searching, chen2021neural}. NAS-Bench-NLP \cite{klyuchnikov2020bench} is a benchmark for neural architecture search which is a collection of 14,322 recurrent architectures; all of the architectures were trained on the PTB dataset. We took 90 randomly selected architectures and compared word embeddings by RTD: each architecture contains 400-dimensional embeddings of 10,000 words. Then, we evaluated all the pairwise similarities between embeddings\footnote{to speedup computation, we averaged the metrics for 10 random batches of 100 word embeddings. The average relative std. dev. of RTD was $8\%$.} from the architectures and visualized them via multi-dimensional scaling, see Fig. \ref{fig:nas-bench-nlp}, where color depicts a log. perplexity. According to common sense, architectures having similar embeddings have a similar log. perplexity. Also, we checked that RTD is approximately a metric for this particular case since it satisfies the triangle inequality for 97\% of triplets of architectures from NAS-Bench-NLP.

\subsection{Experiments with convolutional neural networks} 
\label{sec:cnn} 
To demonstrate the abilities of RTD to work with image representations, we train ResNet-20 \citep{he2016deep} and VGG-11 \citep{simonyan2014very} networks on CIFAR \citep{krizhevsky2009learning} datasets. In experiments, we compare RTD with CKA and disagreement of predictions. For a more intuitive comparison, we consider $1-$CKA instead of CKA. As a measure of the difference in predictions, we use Disagreement \citep{kuncheva2003measures, Wen2020BatchEnsembleAA}, the fraction of mismatched predictions calculated as $\frac{1}{N} \sum_{n=1}^{N}\left[f_{\theta_1}({x}_{n}) \neq f_{\theta_2}({x}_{n})\right]$ , where $f_{\theta}(x)$ denotes the class label predicted by the network for input $x$. As discussed in \citep{fort2019deep}, the lower the accuracy of predictions, the higher its potential mismatch due to the possibility of the wrong answers being random, and then we normalize the Disagreement by $(1 - a)$, where $a$ is the mean accuracy of the predictions. To calculate the final metrics, we averaged the values for five random batches of 500 representations from the test dataset. 

\subsubsection{Training dynamics}

\begin{table}
    \centering
    \vskip-0.05in
    \caption{The correlation of metrics with Disagreement in the training dynamics experiment}
    \begin{tabular}{lll}
              & RTD  & $1-$CKA \\ \hline
    VGG-11    & \textbf{0.976 $\pm$ 0.003} & 0.818 $\pm$ 0.010    \\ 
    ResNet-20 & \textbf{0.971 $\pm$ 0.001} & 0.924 $\pm$ 0.008 
    \end{tabular}
    \label{tbl:evolution}
    \vskip-0.2in
\end{table}

In the first experiment, we analyze the training dynamics of neural networks. On each epoch, we collect the outputs of the convolutional part that extract the representations. To compare dynamics properly, we scaled the metrics by their maximum value. Fig. \ref{fig:evolution} shows the dynamics of the differences with the final representations.
The results coincide with the intuition: the representations on each epoch become more similar to the final one. Moreover, RTD demonstrates the same behavior as disagreement of predictions. RTD better correlates with the Disagreement, see Table \ref{tbl:evolution}.

\subsubsection{Layers}

\begin{figure}
    \centering
    \includegraphics[width=0.35\textwidth]{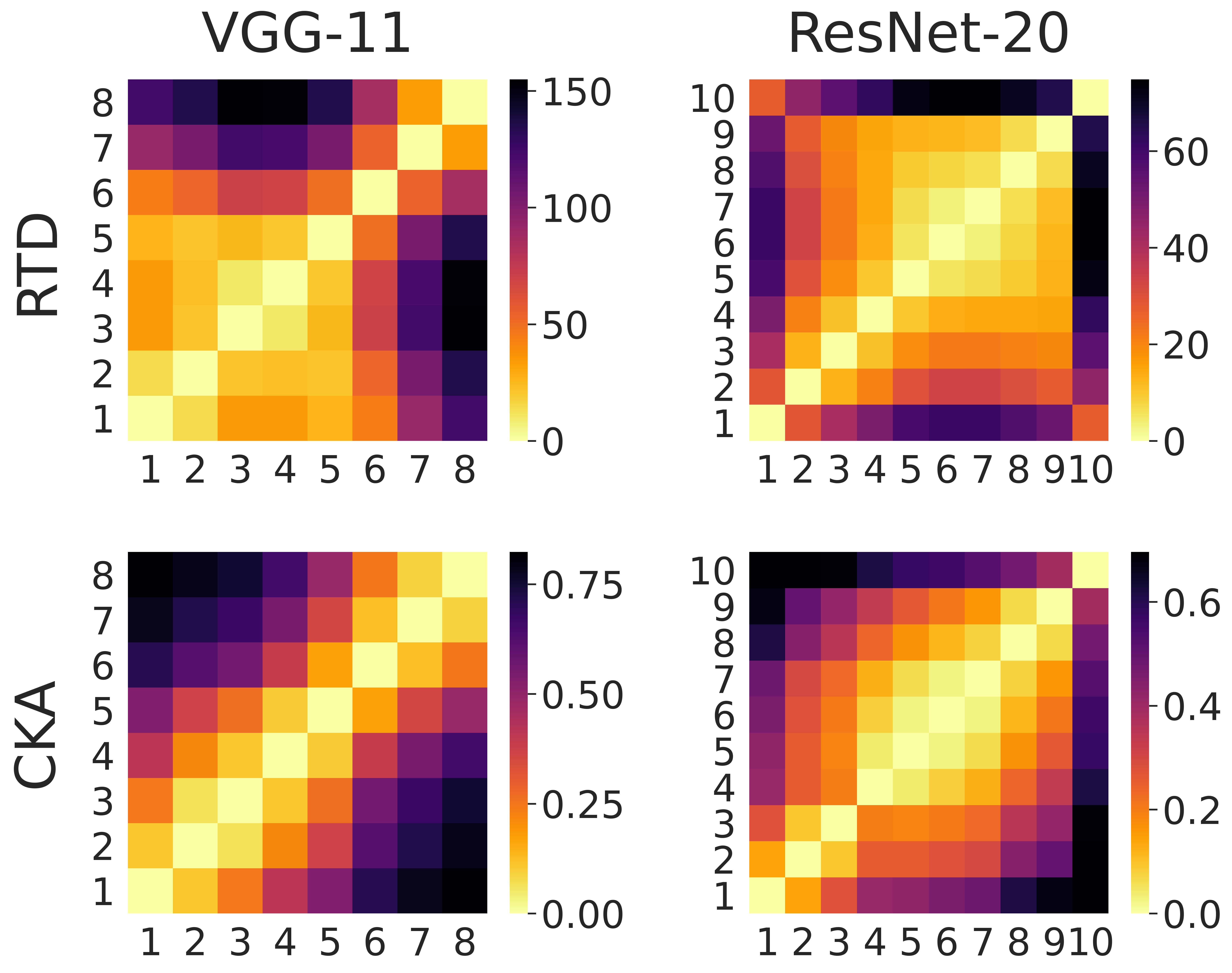}
    \caption{The representation differences between the layer blocks within trained networks. The columns correspond to the architecture, and the rows,  to the metric.}
    \label{fig:layers}
    \vskip-0.2in
\end{figure}

In the next experiment, we compare the outputs of layer blocks within the trained network. For VGG-11, the block has the form  Conv$\to$BN$\to$Activation$\to$(Pooling), and for ResNet-20, we take the output of the first Conv$\to$BN$\to$Activation block, and then the outputs of each residual block. 
In Figure \ref{fig:layers}, we see that both RTD and $1-$CKA show similar results, including the slight difference between adjacent layers. We see that both metrics reveal the significant changes in the outputs of the ResNet-20 last block. In Figure \ref{fig:imagenet-layers}, we performed similar experiment with ResNet-50 and ConvNeXt-tiny \cite{liu2022convnet} architectures pre-trained on ImageNet-1k dataset \cite{deng2009imagenet}.  

\subsubsection{Data Distribution Shift}

\begin{figure*}[t]
\centering
\begin{subfigure}{.24\textwidth}
  \centering
  \includegraphics[width=1\linewidth]{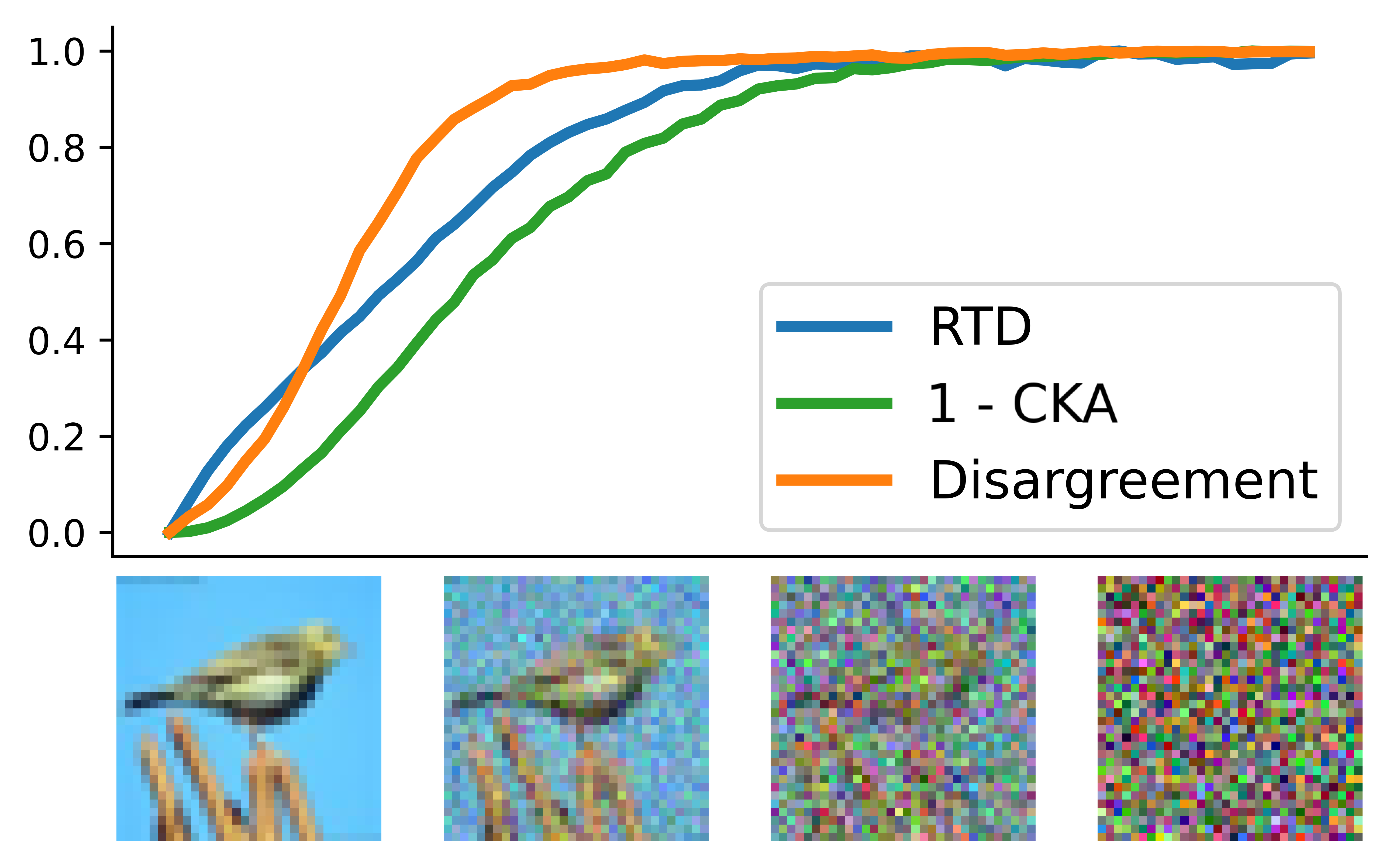}
  \caption{Noise}
  \label{fig:noise}
\end{subfigure}%
\begin{subfigure}{.24\textwidth}
  \centering
  \includegraphics[width=1\linewidth]{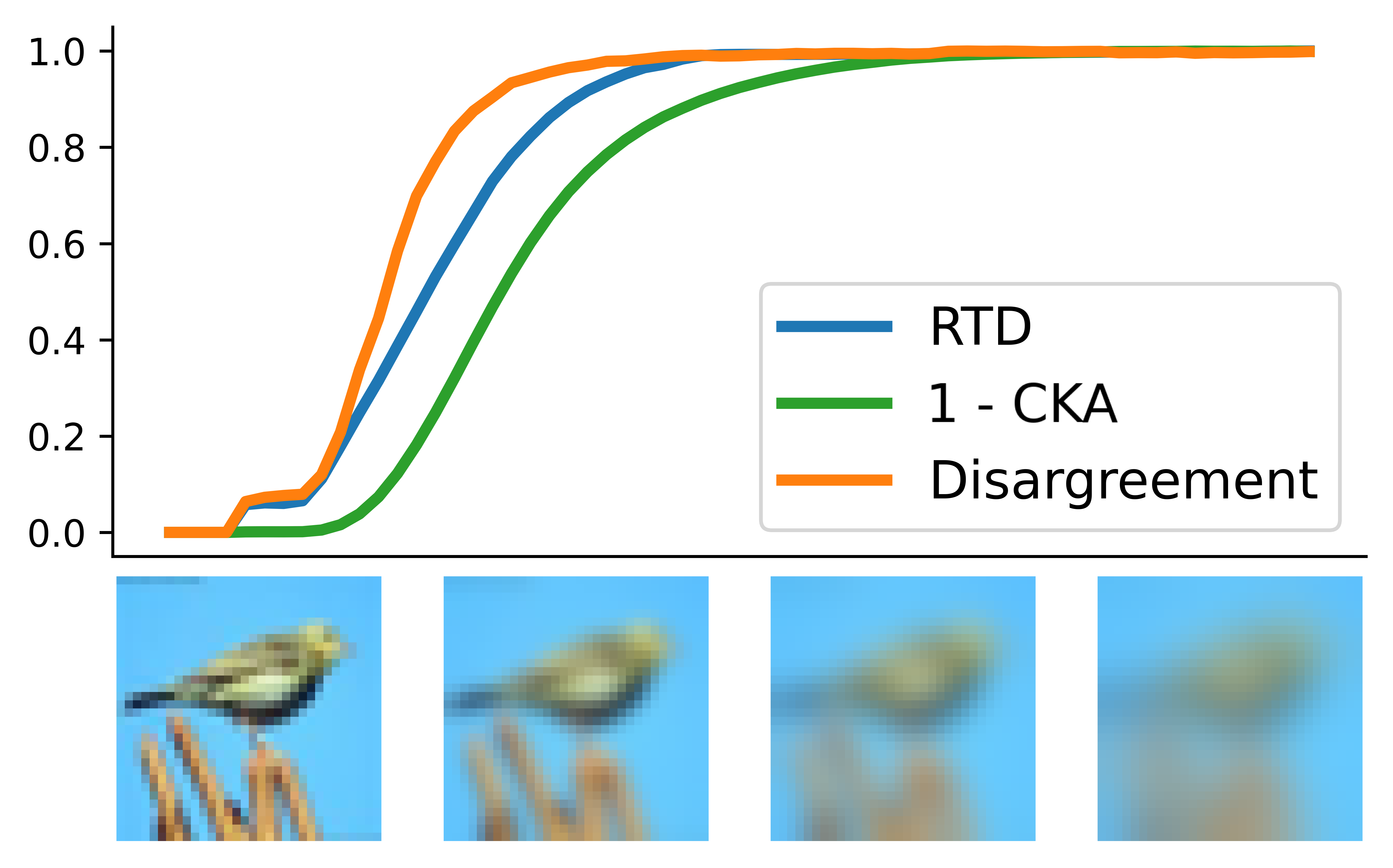}
  \caption{Gaussian blur}
  \label{fig:blur}
\end{subfigure}
\begin{subfigure}{.24\textwidth}
  \centering
  \includegraphics[width=1\linewidth]{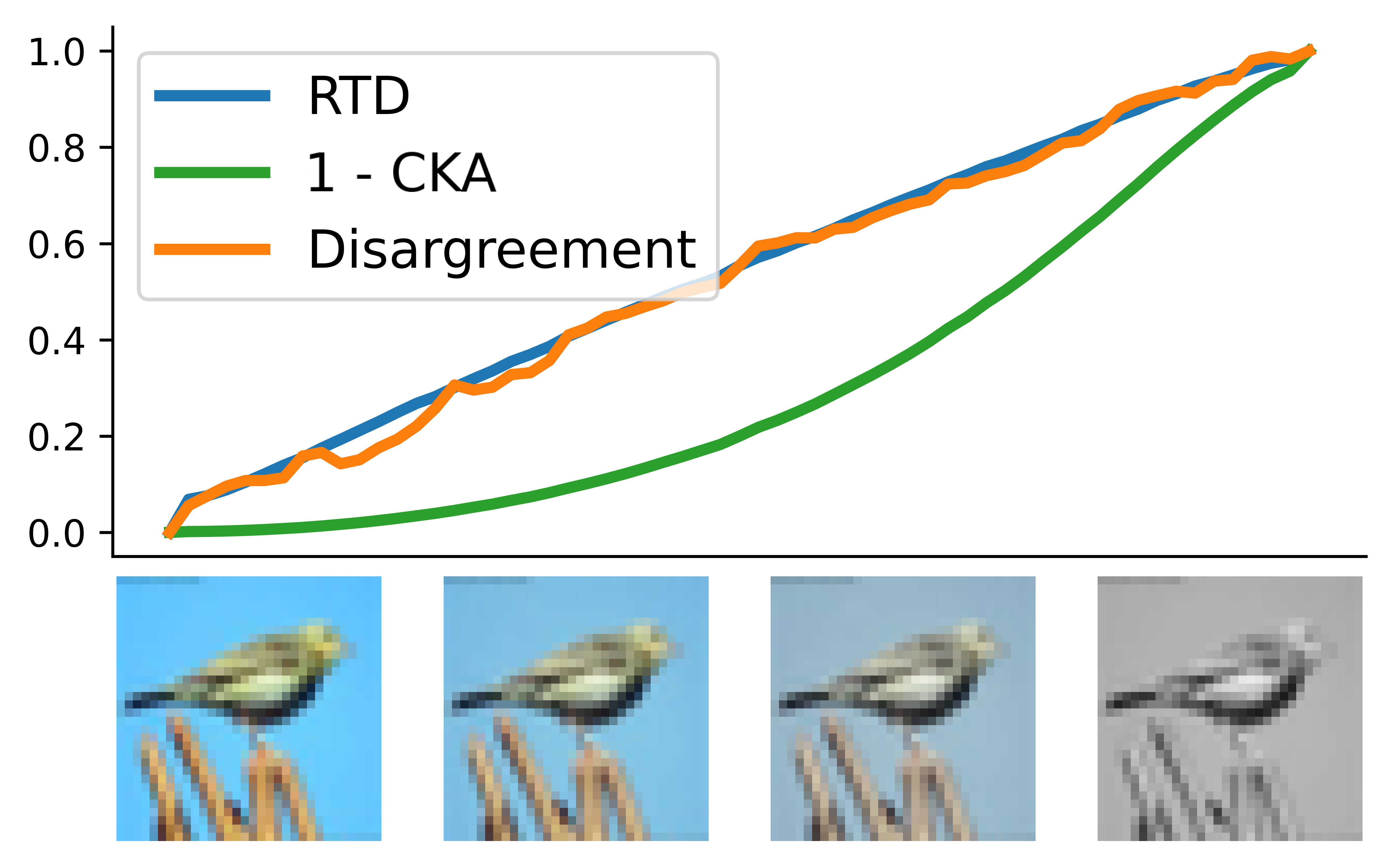}
  \caption{Grayscale}
  \label{fig:grayscale}
\end{subfigure}
\begin{subfigure}{.24\textwidth}
  \centering
  \includegraphics[width=1\linewidth]{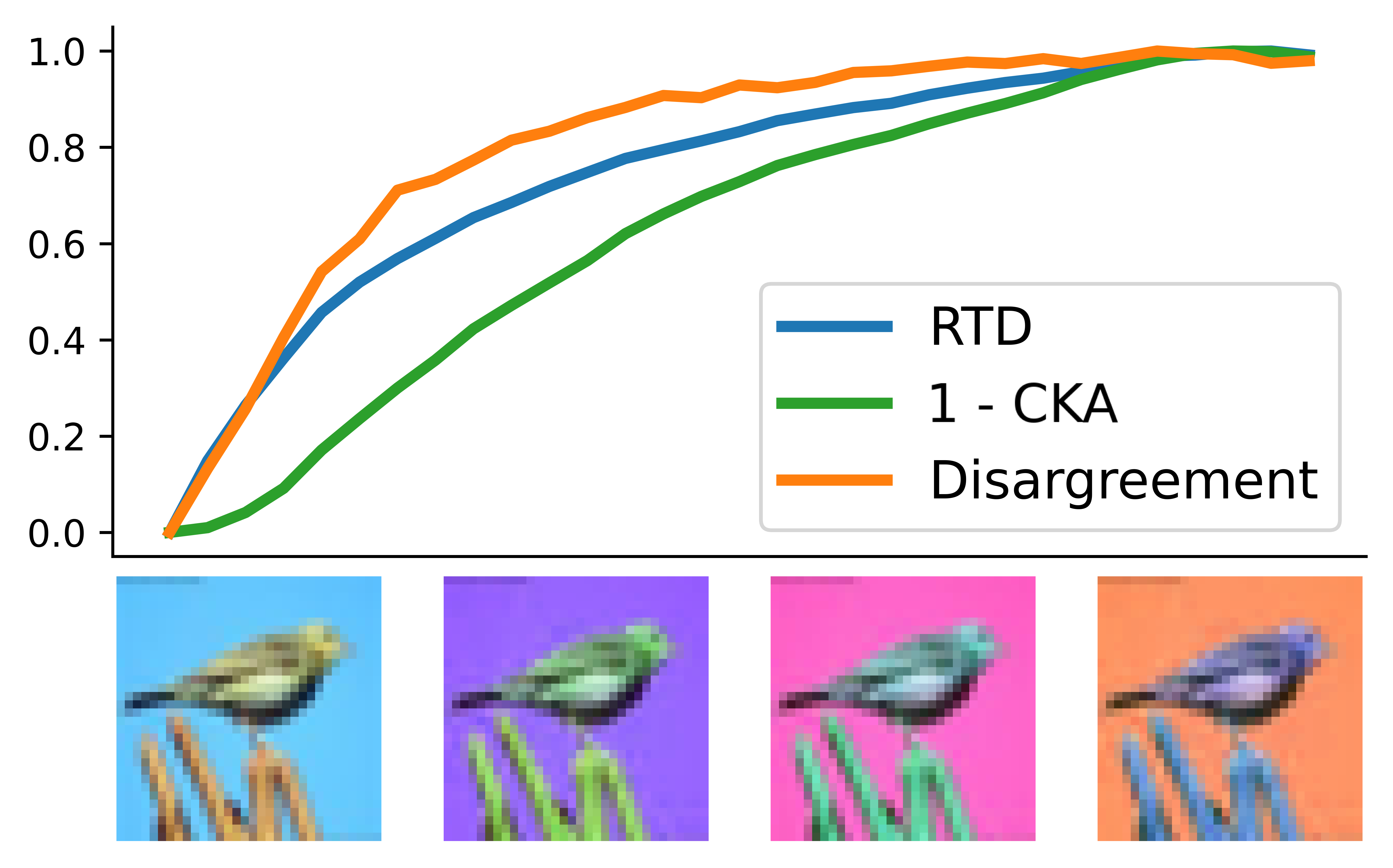}
  \caption{Hue}
  \label{fig:hue}
\end{subfigure}
\caption{Analysis of ResNet-20 representations under different data distribution shifts. The dynamics of scaled metrics with the monotonic transformations of images.}
\label{fig:shift}
\end{figure*}

Here, we apply the data distribution shift to test the RTD. As a shift, we consider different image transformations: noising, blurring, grayscaling, and hue changing. For each transformation, we analyze the metric dynamics as the strength of a transformation increases. Figure \ref{fig:shift} confirms our sanity check of the monotony of RTD and other metrics with respect to data distribution shift. Moreover, Table \ref{tbl:shift} shows that RTD has a higher correlation with disagreement of predictions.

\begin{table}[]
\centering
  \caption{Analysis of ResNet-20 representations under different data distribution shifts. The correlation of metrics with Disagreement.}
    \begin{tabular}{lll}
                  & RTD                                  & $1-$CKA    \\ \hline
    Noise         & \textbf{0.966 $\pm$ 0.001}      & 0.927 $\pm$ 0.006      \\
    Gaussian blur & \textbf{0.982 $\pm$ 0.004}      & 0.913 $\pm$ 0.011      \\
    Grayscale     & \textbf{0.990 $\pm$ 0.004}      & 0.928 $\pm$ 0.040      \\
    Hue           & \textbf{0.978 $\pm$ 0.008}     & 0.927 $\pm$ 0.017     
    \end{tabular}
    \label{tbl:shift}
\end{table}

\subsubsection{Ensembles}

It is known that an ensemble of neural networks performs better than a single network and can estimate the uncertainty of the predictions. It is shown in \citep{lee2015m, opitz1996generating} that the diverse ensembles work better. Thus, measuring ensembles' diversity is important. The disagreement is a good example of such a measure. To show that RTD can measure the diversity as well as disagreement, we learn two types of ensembles: the classical ensemble, when we learn the networks from different random initializations, and the Fast Geometric Ensemble (FGE) \citep{garipov2018loss}, which is known to have lower diversity. We learn four models for each type of ensemble and average the metrics among all pairs. The results in Table \ref{tbl:ensemble} confirm that RTD is capable of measuring the diversity on the same scale as the disagreement of predictions.

\begin{table}[tbp]
\centering
\caption{The averaged metric among all pairs of ensemble members with a ResNet-20 architecture, and the relative difference between the types of ensemble.}
\begin{tabular}{l@{\hspace{0.6\tabcolsep}} c@{\hspace{0.6\tabcolsep}} c@{\hspace{0.6\tabcolsep}}c}
             & Class. Ensemble & FGE               & Diff. \% \\ \hline
RTD          & 15.27 $\pm$ 0.12    & 10.45 $\pm$ 0.32  & \textbf{31.6}         \\ 
$1-$CKA          & 0.094 $\pm$ 0.02   & 0.033 $\pm$ 0.003 & 64.9         \\ 
Disagreement & 0.915 $\pm$ 0.05   & 0.607 $\pm$ 0.03  & \textbf{33.6}          \\ 
\end{tabular}
\label{tbl:ensemble}
\end{table}

\subsubsection{Transfer learning}

\begin{table}[tbp]
    \centering
    \caption{The correlation of metrics with Disagreement in the transfer learning experiment}
    \begin{tabular}{lll}
              & RTD  & $1-$CKA \\ \hline
    CIFAR-100    & \textbf{0.98 $\pm$ 0.01} & 0.93 $\pm$ 0.02  \\ 
    CIFAR-10 & \textbf{0.91 $\pm$ 0.01} & 0.89 $\pm$ 0.02
    \end{tabular}
    \label{tbl:tranfer}
\end{table}

\begin{figure*}[t]
\centering
\begin{subfigure}{.45\textwidth}
  \centering
  \includegraphics[width=1\linewidth]{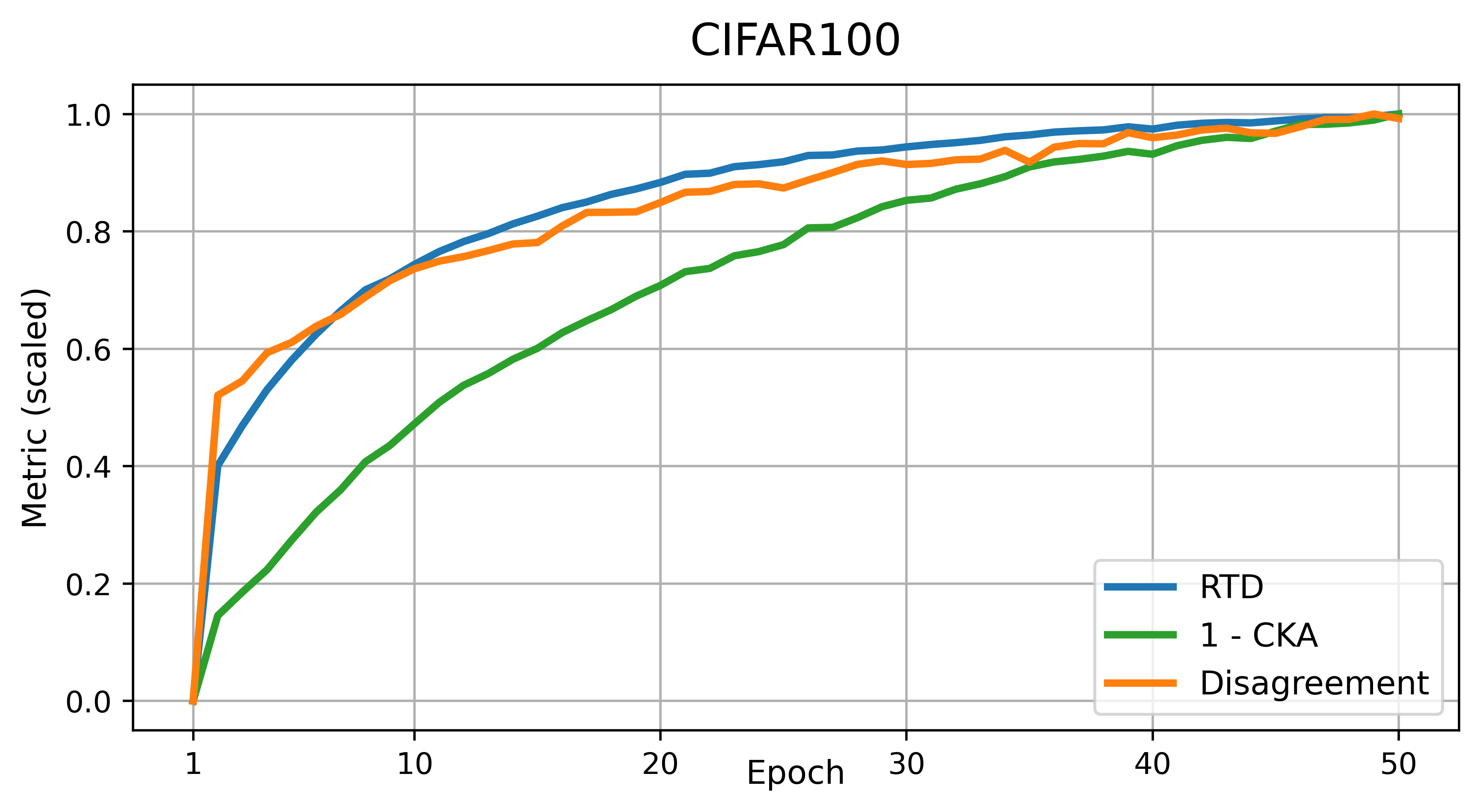}
\end{subfigure}%
\begin{subfigure}{.45\textwidth}
  \centering
  \includegraphics[width=1\linewidth]{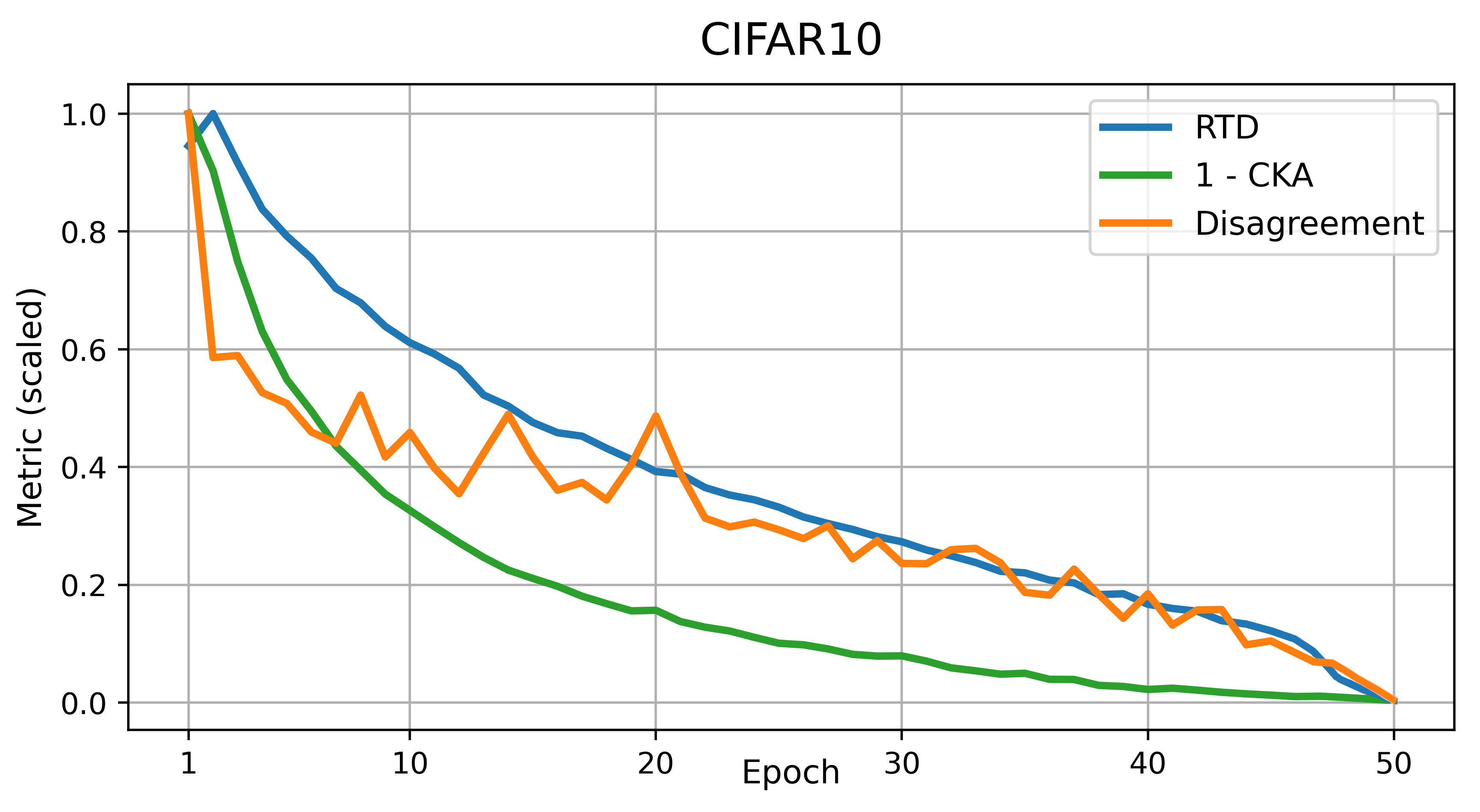}
\end{subfigure}
\caption{Scaled metrics demonstrating the difference between representations of CIFAR-100 and CIFAR-10 datasets during fine-tune process.}
\label{fig:transfer}
\end{figure*}

Another possible application is the measure of changes in representations after transferring the pre-trained model to a new task. In this experiment, we conduct the transfer learning from CIFAR-100 to the CIFAR-10 dataset. We make full fine-tuning with the small learning rate for the convolutional part. In Fig. \ref{fig:transfer}, we demonstrate the dynamics for both dataset representations. The results again coincide with the intuition about the difference during the learning steps, and here RTD has also a high correlation with Disagreement, see Table \ref{tbl:tranfer}. Also, we note that RTD can be applied to the continual learning task, where catastrophic forgetting appears, and thus it is crucial to track the changes in network representations.

\subsection{Additional experiments}
We describe how RTD can be used to evaluate a disentanglement of generative models in Appendix \ref{sec:disent}. Comparisons of BigGAN's internal representations by RTD agree with those of images by FID, see Appendix \ref{sec:biggan}.

\section{Conclusions}

In this paper, we have proposed a topologically-inspired approach to compare neural network representations.
The most widely used methods for this problem are statistical: Canonical Correlation Analysis (CCA) and Centered Kernel Alignment (CKA). But the problem itself is a geometric one: the comparison of two neural representations of the same objects is de-facto the comparison of two points clouds from different spaces.
The natural way is to compare their geometrical and topological features with due account of their localization\,---\, that is exactly what was done by the R-Cross-Barcode and RTD. 
We demonstrated that RTD coincides with the natural assessment of representations similarity.
We used the RTD to gain insights into neural network representations in computer vision and NLP domains for various problems: training dynamics analysis, data distribution shift, transfer learning, ensemble learning, and disentanglement assessment.

RTD correlates strikingly well with the disagreement of models' predictions; this is an intriguing topic for further research.   
Finally, R-Cross-Barcode and RTD are general tools that are not limited to the comparison of representations only. They could be applied to other problems involving comparison of two point clouds with one-to-one correspondence, for example, in 3D computer vision.

\textbf{Acknowledgements}.
The work was supported by the Analytical center under the RF Government (subsidy agreement 000000D730321P5Q0002, Grant No. 70-2021-00145 02.11.2021).


\bibliography{references}
\bibliographystyle{icml2022}

\newpage
\appendix
\onecolumn
\section{Background on Simplicial Complexes. Barcodes}
\label{app:background}

The  simplicial complex is a combinatorial object that can be thought of as a higher-dimensional generalization of a graph.

A simplex is defined via the set of its vertices. Given a finite set $V$, a $k-$simplex is a finite $(k+1)-$element subset in $V$. 
Simplicial complex $S$ is a collection of  $k-$simplexes, $k\geq 0$, which satisfies the natural condition that for each $\sigma \in S$,  $\sigma' \subset \sigma$ implies $\sigma'\in S$. A simplicial complex consisting only of $0-$ and $1-$simplexes is a graph.

Denote via $C_k(S)$ the vector space over the field $\mathbb{Z}/2\mathbb{Z}=\{0,1\}$ whose basis elements are $k-$simplexes from $S$. The boundary linear operator $\partial_k:C_k(S)\to C_{k-1}(S)$ is defined on $\sigma=\{A_0,\ldots,A_{k}\}$ as
\begin{equation*}
    \partial_k \sigma=\sum_{j=0}^k \{A_0,\ldots,A_{j-1},A_{j+1},\ldots,A_{k}\}.
\end{equation*}
The $k$th \textbf{homology} group $H_k(S)$ is  the factor vector space  $\ker\partial_k/\operatorname{im}\partial_{k+1}$. The elements $c\in \ker\partial_k$ are called cycles. The elements of  $H_k(S)$ represent various $k-$dimensional topological features in $S$. A basis in $H_k(S)$ corresponds to a set of basic topological features. 

For example,  the vector space $H_0$ has the basis whose elements are in one-to-one correspondence with equivalence classes of vertices, connected by paths of $1-$simplices (edges), i.e. with connected components of $S$. The basis elements of the vector space $H_1$ correspond to basic equivalence classes of nontrivial  closed paths of $1-$simplices. Two closed paths, also named $1-$cycles, are equivalent if they are connected by a chain of modifications by boundaries of triangles ($2-$simplices). 

A map $S_1\to S_2$, e.g. $\mathcal{G}^{{w}\leq \alpha}\to \mathcal{G}^{\min(w,\tilde{w})\leq \alpha}$ see section \ref{sec:Topfeatures}, defines the maps $H_k(S_1)\to H_k(S_2)$. The kernel of the linear map
$H_0(S_1)\to H_0(S_2)$  is spanned by the pairs of $S_1$ clusters merged together in $S_2$. The cokernel of the linear map  $H_1(S_1)\to H_1(S_2)$ consists of $1-$cycles in $S_2$ which are not from $S_1$, i.e. it consists of  equivalence classes of closed paths in $S_2$, which cannot be modified by boundaries of triangles into images of $1-$cycles from $S_1$.

In applications, the simplicial complexes are often built via consequential adding of simplexes one after another in increasing order of some numerical characteristics. Mathematically this corresponds to a filtration on the simplicial complex. It is defined as a family of simplicial complexes $S_\alpha$, indexed by a finite set of real numbers, with nested collections of simplexes: for $\alpha_1 < \alpha_2$ all simplexes of $S_{\alpha_1}$ are also in $S_{\alpha_2}$. An example of a filtered simplicial complex is the Vietoris-Rips simplicial complex from Section \ref{subs:RCrB}.

The inclusions $S_{\alpha}\subseteq S_{\beta}$ induce the maps on homology $H_k(S_{\alpha})\to H_k(S_{\beta})$.
The evolution of cycles across the nested family of simplicial complexes $S_{\alpha_i}$ is described by the principal persistent homology theorem \citep{B94,zomorodian2001computing,viterbo:2011}, according to which for each dimension there exists a choice of a set of basic topological features across all nested simplicial complexes $S_{\alpha}$ so that each basic feature $c$ appears in $H_k(S_{\alpha})$ at specific time $\alpha=b_c$ and disappears at specific time $\alpha=d_c$. The barcode of the filtered complex is the record of the appearance, or ``birth'' time, and the disappearance, or ``death'' time, of all these basic topological features. 

\subsection{Exact sequence and topological features}
A sequence of vector spaces and linear maps\begin{equation}
A_5\xrightarrow{r_{4}}   A_4\xrightarrow{r_{3}}A_3\xrightarrow{r_{2}}A_2 \xrightarrow{r_{1}}A_1\label{eq:longseq2}
\end{equation}
is exact at $A_j$ if the kernel of the linear map $r_{j-1}$ coincides with the image of the previous map $r_{j}$. 
\begin{proposition}
If the sequence (\ref{eq:longseq2}) is exact at $A_2,A_3,A_4$ then $A_3\simeq \operatorname{Ker}(r_1) \oplus \operatorname{Coker}(r_4)$.
\end{proposition}\vspace{-6mm}
\begin{proof}
Since
$A/\operatorname{Ker}(r)\simeq \operatorname{Image}(r)$
for any linear map $r:A\to A'$, therefore $A_3\simeq \operatorname{Image}(r_2)\oplus \operatorname{Ker}(r_2)$. If the sequence is exact at $A_2$, then $\operatorname{Image}(r_2)\simeq \operatorname{Ker}(r_1)$. Exactness at $A_3$, $A_4$ gives $\operatorname{Ker}(r_2)\simeq \operatorname{Image}(r_3)$,  $\operatorname{Ker}(r_3)\simeq \operatorname{Image}(r_4)$. Then $\operatorname{Image}(r_3) \simeq  A_4/\operatorname{Ker}(r_3)$ imply that $\operatorname{Ker}(r_2)\simeq A_4/\operatorname{Image}(r_4)$, which equals $\operatorname{Coker}(r_4)$, the cokernel of the linear map $r_4$.
Hence $A_3\simeq \operatorname{Ker}(r_1) \oplus \operatorname{Coker}(r_4)$. \end{proof}

Therefore the exact sequence from Theorem \ref{thr:basic} implies that the calculation of the topological features from Section \ref{sec:Topfeatures} for all $\alpha$ is reduced to the calculation of
$H_1(R_\alpha(\hat{\mathcal{G}}^{w,\tilde{w}}))$ for all $\alpha$, i.e. to the calculation of  \emph{R-Cross-Barcode$_1(P,\tilde{P})$}.

\subsection{Construction of R-Cross-Barcode}
Here we gather some intuition behind the construction of the graph $\hat{\mathcal{G}}^{w,\tilde{w}}$ and the \emph{R-Cross-Barcode}.

The Vietoris-Rips complex $R_\alpha(\mathcal{G}^{\min(w,\tilde{w})})$ is the union of simplexes whose edges connect data points with distance less than $\alpha$ in at least one of representations $\cal{P},\tilde{\cal{P}}$. An inclusion of simple simplicial complexes $S\subset R$  is an equivalence in homotopy category, if and only if the induced map on homology is an isomorphism   \citep{whitehead2012elements}.
The 
 maps on homology
induced by the inclusions of filtered simplicial complexes 
\begin{equation}\label{wu_to_min}
   R_\alpha(\mathcal{G}^w)\subseteq R_\alpha(\mathcal{G}^{\min(w,\tilde{w})}),\,\,  R_\alpha(\mathcal{G}^{\tilde{w}}) \subseteq R_\alpha(\mathcal{G}^{\min(w,\tilde{w})})
\end{equation}
should therefore be as close as possible to isomorphisms, in order that the approximations at threshold $\alpha$ to the manifolds $M_{\mathcal{P}}$ and  $M_{\tilde{\mathcal{P}}}$ have essentially the same geometric features located at the same places. 
It follows from the exact sequence from Theorem \ref{thr:basic} that the R-Cross-Barcode$_*(P,\tilde{P})$ is exactly the list of topological features describing the failure of the maps induced on homology by inclusions (\ref{wu_to_min}) to be isomorphisms.

\label{propRcross}

Introduce the
weighted graph 
$\hat{\mathcal{G}}^{w}$ with doubled set of vertices and with the edge weights defined as follows. We fix the numbering of vertices $\text{Vert}(\mathcal{G})=\{A_1,\ldots,A_N\}$.
Let us add the extra vertex $A'$ together with $A$ 
to $\hat{\mathcal{G}}^{w}$ for each vertex $A\in \text{Vert}(\mathcal{G})$, plus the two additional vertexes $O,O'$, and define the distance-like edge weights in $\hat{\mathcal{G}}^{w}$ as:
\begin{gather}
    d_{A_iA_j}=d_{A_iA'_j}=w_{A_iA_j} , \nonumber\\ d_{A'_iA'_j}=d_{A_iA'_i}=d_{O'A'_i}=d_{OA_i}=0,\,\, d_{A'_jA_i}=d_{O'A_i}=d_{OA'_i}=d_{OO'}=+\infty \label{eq:matr1}
\end{gather}
where $i<j$ and $O,O'\in \text{Vert}(\hat{\mathcal{G}}^w)$ are the two additional vertexes.



The suspension $C[-1]$ of chain complex $C$ denotes the same chain complex with degree shifted by $1$, $C[-1]_n=C_{n-1}$, so that the the $n$th chains of $R_\alpha(\mathcal{G}^{w})[-1]$ are linear combinations of  $(n-1)-$dimensional simplexes from  $R_\alpha(\mathcal{G}^{w})$. We denote via $A_{i_1}\ldots A_{i_{n}}[-1]$ the element from   $C_{n}(R_\alpha(\mathcal{G}^{w})[-1])$   corresponding to the simplex $A_{i_1}\ldots A_{i_{n}}$.

A chain map $f$ between two chain complexes $(C,d_{C})$ and $(B,d_{B})$ is a sequence of linear maps $f_{n}:C_{n}\rightarrow B_{n}$  that commutes with the boundary operators: $d_{B,n}\circ f_{n}=f_{n-1}\circ d_{C,n}$.
The cone of a chain map $f$ is the chain complex
$\operatorname{Cone}(f)=C[-1]\oplus B$ with differential
$d_{\operatorname{Cone}(f)}={\begin{pmatrix}d_{C[-1]}&0\\f[-1]&d_{B}\end{pmatrix}}$. 
A homotopy equivalence is a pair of chain maps $f:C\to B$, $g:B\to C$, and a pair of  maps $h_{C,n}:C_n\to C_{n+1}$, $h_{B,n}:B_n\to B_{n+1}$, such that $g\circ f=\operatorname{Id}+[h_C,d_C]$ and $f\circ g=\operatorname{Id}+[h_B,d_B]$. We assume that the Vietoris-Rips complexes are augmented with $C_{-1}=\mathbb{Z}/2\mathbb{Z}$ and 
$\partial_0 \{A_i\}=1$.

The proof of the exact homology sequence from Theorem \ref{thr:basic} follows from the following two propositions.

\begin{proposition}
There are homotopy equivalences of chain complexes:
 \begin{gather}
 R_\alpha(\mathcal{G}^{w})[-1]\sim R_\alpha(\hat{\mathcal{G}}^w) \label{homeq1}\\
     \operatorname{Cone}\left( R_\alpha(\mathcal{G}^{w})\to  R_\alpha(\mathcal{G}^{\min(w,\tilde{w})})\right)\sim R_\alpha(\hat{\mathcal{G}}^{w,\tilde{w}}). \label{homeq2}
 \end{gather}
\end{proposition}
\begin{proof}
The simplexes of the chain complex $R_\alpha(\hat{\mathcal{G}}^w)$ are of four types: $A_{i_1}\ldots A_{i_k}A'_{i_k}\ldots A'_{i_{n}}$, $A_{i_1}\ldots A_{i_k}A'_{i_{k+1}}\ldots A'_{i_n}$, $OA_{i_1}\ldots A_{i_n}$ and $O'A'_{i_1}\ldots A'_{i_n}$ where $A_{i_k}\in \text{Vert}(\mathcal{G})$, $i_0<\ldots<i_k<i_{k+1}<\ldots<i_n$, with edge weights satisfying $w_{A_{i_r}A_{i_s}}<\alpha$ for $r\leq k$. Define the map  
$\phi: R_\alpha(\mathcal{G}^{w})[-1]\to R_\alpha(\hat{\mathcal{G}}^w)$
\begin{gather}
   \phi:A_{i_1}\ldots A_{i_{n}}[-1] \mapsto OA_{i_1}\ldots A_{i_{n}}+O'A'_{i_1}\ldots A'_{i_{n}}+\sum_{k=1}^nA_{i_1}\ldots A_{i_k}A'_{i_k}\ldots A'_{i_{n}}\label{eq:hom1}
\end{gather}  
 The map $\phi$  together with the  map $\tilde{\phi}:R_\alpha(\hat{\mathcal{G}}^w)\to R_\alpha(\mathcal{G}^{w})[-1]$    
\begin{gather}
 \tilde{\phi}:OA_{i_1}\ldots A_{i_{n}}\mapsto A_{i_1}\ldots A_{i_n}[-1], \,\,
\tilde{\phi}(\Delta)=0 \text{ for any other simplex }\Delta,
\end{gather}
gives a homotopy equivalence,  $\tilde{\phi}\circ\phi=\operatorname{Id}$, $\phi\circ\tilde{\phi}=\operatorname{Id}+[h,\partial]$, where the homotopy $h$ is given by
\begin{gather}
    h: A_{i_1} \ldots A_{i_{k}}A'_{i_{k+1}}\ldots A'_{i_n}\mapsto \sum_{l=1}^kA_{i_1} \ldots A_{i_{l}} A'_{i_{l}}\ldots A'_{i_n}+O'A'_{i_1}\ldots A'_{i_n} \\  h:A'_{i_1} \ldots A'_{i_n}\mapsto O'A'_{i_1}\ldots A'_{i_n}, \,\, 
h(\Delta)=0 \text{ for any other simplex }\Delta.\nonumber
\end{gather}
Simplexes of the chain complex $R_\alpha(\hat{\mathcal{G}}^{w,\tilde{w}})$ are of three types. The first type: $A_{i_1}\ldots A_{i_k}A'_{i_k}\ldots A'_{i_{n}}$  with edge weights satisfying $w_{A_{i_r}A_{i_s}}<\alpha$ for $r\leq k$ and $\min(w_{A_{i_r}A_{i_s}},\tilde{w}_{A_{i_r}A_{i_s}})<\alpha$ for $r,s>k$;
the second type: $A_{i_1}\ldots A_{i_{k-1}}A'_{i_{k}}\ldots A'_{i_n}$ with edge weights satisfying $w_{A_{i_r}A_{i_s}}<\alpha$ for  $r<k$, and $\min(w_{A_{i_r}A_{i_s}},\tilde{w}_{A_{i_r}A_{i_s}})<\alpha$ for $r,s\geq k$;
and the third type: $OA_{i_1}\ldots A_{i_n}$ with edge weights satisfying $w_{A_{i_r}A_{i_s}}<\alpha$ for all $r,s$. Define the map  
$\psi: \operatorname{Cone}\left( R_\alpha(\mathcal{G}^{w})\to  R_\alpha(\mathcal{G}^{\min(w,\tilde{w})})\right)\to R_\alpha(\hat{\mathcal{G}}^{w,\tilde{w}})$
\begin{gather}
   \psi:A_{i_1}\ldots A_{i_{n}}[-1] \mapsto OA_{i_1}\ldots A_{i_{n}}+\sum_{k=1}^nA_{i_1}\ldots A_{i_k}A'_{i_k}\ldots A'_{i_{n}}
\end{gather} for $A_{i_1}\ldots A_{i_{n}}[-1]\in R_\alpha(\mathcal{G}^{w})[-1]$,
\begin{gather}
   \psi:A_{i_1}\ldots A_{i_{n}} \mapsto A'_{i_1}\ldots A'_{i_{n}}
\end{gather} for $A_{i_1}\ldots A_{i_{n}}\in R_\alpha(\mathcal{G}^{\min(w,\tilde{w})})$.
 The map $\psi$  together with the  map $\tilde{\psi}:R_\alpha(\hat{\mathcal{G}}^{w,\tilde{w}}) \to\operatorname{Cone}\left( R_\alpha(\mathcal{G}^{w})\to  R_\alpha(\mathcal{G}^{\min(w,\tilde{w})})\right)$
 \begin{gather}
 \tilde{\psi}:OA_{i_1}\ldots A_{i_{n}}\mapsto A_{i_1}\ldots A_{i_n}[-1], \,\,A_{i_1}\ldots A_{i_{n}}[-1]\in R_\alpha(\mathcal{G}^{w})[-1],\nonumber\\
 A'_{i_1}\ldots A'_{i_{n}}\mapsto A_{i_1}\ldots A_{i_n}, \,\,A_{i_1}\ldots A_{i_{n}}\in R_\alpha(\mathcal{G}^{\min(w,\tilde{w})}),\\
\tilde{\psi}(\Delta)=0 \text{ for any other simplex }\Delta,\nonumber
\end{gather}
gives a homotopy equivalence,  $\tilde{\psi}\circ\psi=\operatorname{Id}$, $\psi\circ\tilde{\psi}=\operatorname{Id}+[H,\partial]$, where the homotopy $H$ is given by
\begin{gather}
    H: A_{i_1} \ldots A_{i_{k}}A'_{i_{k+1}}\ldots A'_{i_n}\mapsto \sum_{l=1}^kA_{i_1} \ldots A_{i_{l}} A'_{i_{l}}\ldots A'_{i_n},\,\, 1\leq k\leq n \label{eq:hom2}\\   
H(\Delta)=0 \text{ for any other simplex }\Delta.\nonumber
\end{gather} 
\end{proof}



The long exact sequences such as (\ref{eq:longseq}) arise from distinguished triangles in the homotopy category of chain complexes. A distinguished triangle is a diagram isomorphic in this category to a diagram 
$A{\xrightarrow {f}}B\to \operatorname{Cone}(f)\to A[-1]$.
\begin{proposition} 
 The embeddings of graphs $\mathcal{G}^{w\leq\alpha}\subseteq \mathcal{G}^{\min(w,\tilde{w})\leq\alpha}\subset\hat{\mathcal{G}}^{w,\tilde{w}\leq\alpha} $
give distinguished triangles, see \citep{gelfand2002methods}, in the homotopy category of chain complexes: \begin{equation}
     R_\alpha({\mathcal{G}}^w)\to  R_\alpha(\mathcal{G}^{\min(w,\tilde{w})})\to R_\alpha(\hat{\mathcal{G}}^{w,\tilde{w}})\to R_\alpha({\mathcal{G}}^w)[-1].
     \label{eq:triang1}
\end{equation} 

\end{proposition}
\begin{proof}
Taken together the homotopy equivalences (\ref{eq:hom1})-(\ref{eq:hom2}) define an isomorphism of (\ref{eq:triang1}) with the distinguished triangle \begin{equation}
     R_\alpha({\mathcal{G}}^w)\to  R_\alpha(\mathcal{G}^{\min(w,\tilde{w})})\to \operatorname{Cone}\left( R_\alpha(\mathcal{G}^{w})\to  R_\alpha(\mathcal{G}^{\min(w,\tilde{w})})\right)\to R_\alpha({\mathcal{G}}^w)[-1].
\end{equation} 
\end{proof}
  

\emph{Comparison with Cross-Barcode and Geometry Score.}
The Cross-Barcode from \citep{barannikov2021manifold} compares two data manifolds lying in the same ambient space. It does not use the information that can be provided by a one-to-one correspondence between points of the two data clouds. To compare the locations of topological features the Cross-Barcode from loc.cit. uses instead the proximity information inferred from the pairwise distances between points from different clouds lying in the same ambient space. Geometry score from \citep{khrulkov2018geometry} is based on a comparison of standard barcodes for each cloud and is insensitive to the location of topological features, for example, it does not detect any difference when similar topological features are located geometrically in distant places of the two clouds.

\begin{figure}[h]
    \centering
    \includegraphics[width=0.25\textwidth]{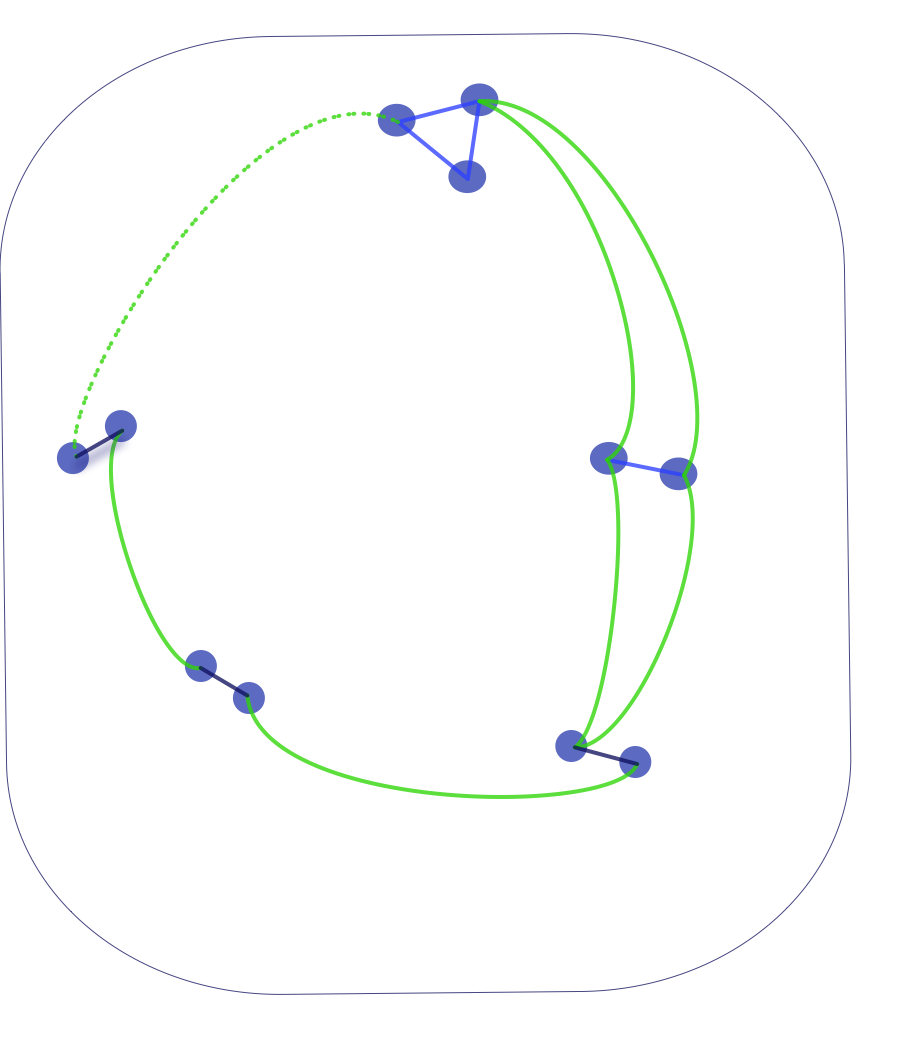}
    \caption{Merging between clusters already connected via a chain of mergings.}
    \label{fig:5clusters}
\end{figure}

\begin{figure}[h!]
    \centering
    \includegraphics[width=0.45\textwidth]{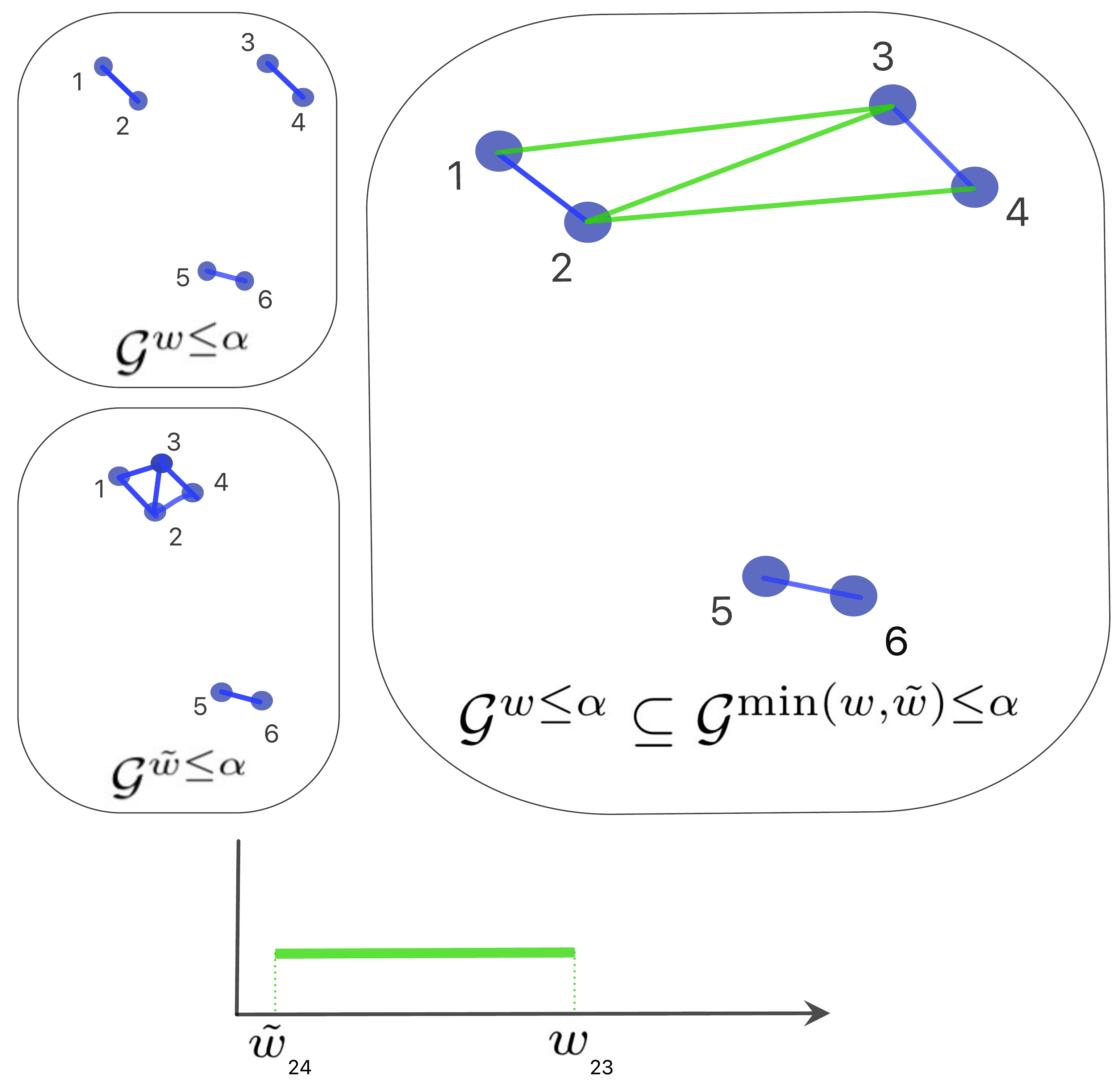}
    \caption{Merging of three clusters into two clusters. Graphs $\mathcal{G}^{{w}\leq \alpha}$, $\mathcal{G}^{\tilde{w}\leq \alpha}$ and  $\mathcal{G}^{\min(w,\tilde{w})\leq \alpha}$ are shown.  Edges of $\mathcal{G}^{\min(w,\tilde{w})\leq \alpha}$ not in $\mathcal{G}^{{w}\leq \alpha}$ are colored in green. In this example there are exactly four different weights $(13),(14),(23),(24)$ in the graphs $\mathcal{G}^{{w}\leq \alpha}$ and  $\mathcal{G}^{\min(w,\tilde{w})\leq \alpha}$.  The unique topological feature in \emph{R-Cross-Barcode$_1(P,\tilde{P})$} in this case  is born at the threshold $\tilde{w}_{24}$ when the difference in the cluster structures of the two graphs arises, as the points $2$ and $4$ are in the same cluster at this threshold in $\mathcal{G}^{\min(w,\tilde{w})}$ and not in $\mathcal{G}^{{w}}$. This feature dies at the threshold ${w}_{23}$ since the clusters containing $2$ and $4$ are merged at this threshold in $\mathcal{G}^{{w}}$.}
    \label{fig:3to2clusters}
\end{figure}

\section{Discussion of CKA}
\label{app:cka_discussion}

Given two series of equal size $x_i\in \mathbb{R}^{n_x}$, $y_i\in \mathbb{R}^{n_y}$, $i=1 \ldots n$ the CKA \citep{kornblith2019similarity} is defined as 
$$
\text{CKA}(K,L)=\frac{\text{HSIC}(K,L)}{\sqrt{\text{HSIC}(K,K)\text{HSIC}(L,L)}}
$$
where $\text{HSIC}(K,L)$ is a Hilbert-Schmidt Independence Criterion \citep{gretton2005measuring},
$K_{i,j}=k(x_i, x_j)$, $L_{i,j}=l(y_i, y_j)$, $L=E-n^{-1}$ where $k(\cdot,\cdot)$, $l(\cdot,\cdot)$ are kernels.
HSIC itself an empirical estimate of the Hilbert-Schmidt norm of the cross-covariance operator. HSIC
is equivalent to maximum mean discrepancy between the joint distribution $P(X,Y)$ and the product of the marginal distributions $P(X)P(Y)$; HSIC = 0 implies independence of $X$ and $Y$ if the associated kernel is universal. 

However, CKA is sometimes applied to measure similarity between representations from different layers of a neural network.
In this case $Y=f(X)$. $X$ and $Y$ are tightly dependent and the joint distribution can always be factorized as $P(X,Y)=P(Y|X)P(X)$. Thus, the application of CKA to the comparison of representation from different layers is questionable.

\section{Details on experiments with synthetic point clouds}
\label{app:synthetic-r-cross-barcodes}

\begin{figure}[h]
    \centering
    \includegraphics[width=\textwidth]{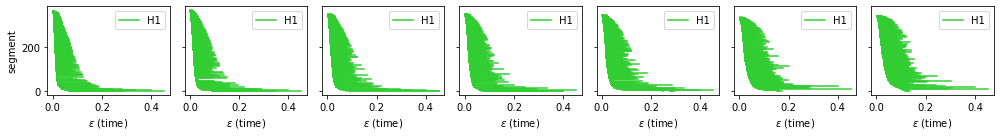}
    \includegraphics[width=\textwidth]{img/clusters_r_cross_barcodes2.png}
    \caption{R-Cross-Barcodes for the ``clusters'' experiments. Top: R-Cross-Barcode$(\tilde{P},P)$, Bottom: R-Cross-Barcode$(P, \tilde{P})$; $\tilde{P}$ is the point cloud having one cluster; ${P}$- 2, 3, 4, 5, 6, 10, 12 clusters.}
    \label{fig:clusters_full1}
\end{figure}

\textbf{Runtime comparison}. Here we present the total wall time of the experiments with synthetic point clouds:

``Clusters experiment'': RTD: 19.7 s, CKA: 0.07 s, IMD: 83 s, SVCCA: 0.03 s.\\
``Rings experiment'': RTD: 144 s, CKA: 0.7 s, IMD: 91 s, SVCCA: 0.6 s.

\section{Details on the ``rings'' experiment}
\label{app:rings}

%
%
\begin{figure*}[th]
\centering
\begin{subfigure}{\textwidth}
\centering
\includegraphics[width=0.8\textwidth]{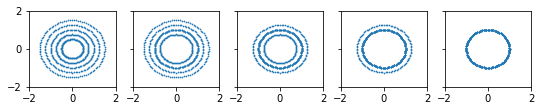}
\caption{Point clouds used in ``rings'' experiment.}
\label{fig:rings_clouds}
\end{subfigure}

%
%

%
%
\begin{subfigure}[t]{\textwidth}
\centering
\includegraphics[width=0.23\textwidth]{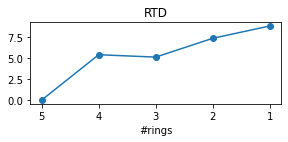}
\includegraphics[width=0.23\textwidth]{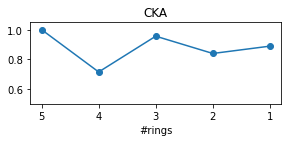}
\includegraphics[width=0.23\textwidth]{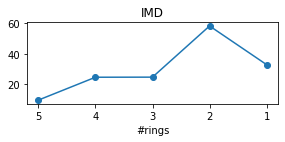}
\includegraphics[width=0.23\textwidth]{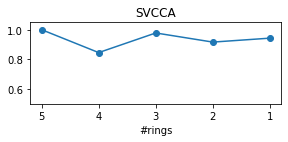}
\caption{Representations' comparison measures. Ideally, the measure should change monotonically with the increase of topological discrepancy.}
\label{fig:rings_metrics}
\end{subfigure}

\caption{RTD perfectly detects changes in topology, while rival measures fail. Five rings are compared with 5,4,3,2,1 rings.}
\label{fig:rings}
\end{figure*}

\begin{figure}[h]
    \centering
    \includegraphics[width=0.7\textwidth]{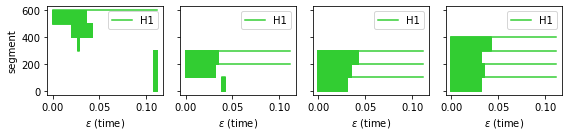}
    \includegraphics[width=0.7\textwidth]{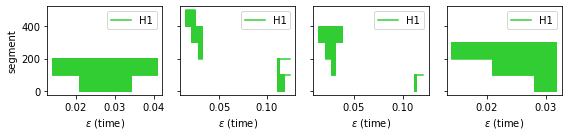}
    \caption{R-Cross-Barcodes for the ``rings'' experiments. Top: R-Cross-Barcode$(P, \tilde{P})$, Bottom: R-Cross-Barcode$(\tilde{P}, P)$. $P$ - is the point cloud having 5 rings, $\tilde{P}$ - 4, 3, 2, 1 rings.}
    \label{fig:clusters_full22}
\end{figure}

\section{Experiment with BigGAN}
\label{sec:biggan}
\begin{figure}
    \centering
    \includegraphics[width=0.3\textwidth]{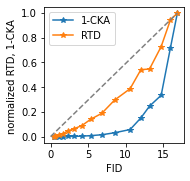}
    \caption{Comparison of normalized RTD, CKA (computed for sets of internal representations) vs. FID (computed for sets of images).}
    \label{fig:biggan}
    \vskip-0.0in
\end{figure}

In this experiment, we applied RTD and CKA for comparison of internal representations in \mbox{BigGAN} \citep{brock2018large} \footnote{we used the pretrained model from\\ \url{https://github.com/lukemelas/pytorch-pretrained-gans}}.

Initially, we generated a set of $k=100$ random latent codes $Z_{0} = \{z_{0,j}\}_{j=1}^k$ and derived sets $Z_1, \ldots, Z_n$ by adding to $Z_0$ a Gaussian noise of increasing strength 
$z_{i,j}= z_{0,j} + \epsilon_{i,j}$, where $\epsilon_{i,j} \sim N(0, \sigma_i)$. 
The noise standard deviation $\sigma_i$ grows from $0.001$ to $0.25$ by a logarithmic scale and the difference between $Z_0$ and $Z_i$ tends to increase when $i$ increases.

Then, we pass sets of latent codes $Z_0, \ldots, Z_n$ together with vector encoding of the ``husky'' class through the \mbox{BigGAN} and save internal representations $R_{i}$ for one of the top layers (results were quite similar for other layers). 
Also, we get sets of images $I_{0}, \ldots, I_{k}$. To compare these sets we used the state-of-the-art measure FID \citep{heusel2017gans} which is often applied for GAN evaluation.

It is natural to assume that the difference between sets of internal representations $R_{0}$ and $R_{i}$ should have a good correlation with the difference between sets of images $I_{0}$ and $I_{i}$.
To check this hypothesis, we calculated RTD($R_0$, $R_i$), CKA($R_0$, $R_i$) and compared them with FID($I_0$, $I_i$), for $i = 1, \ldots, n$. Figure \ref{fig:biggan} shows the results. We conclude that RTD enjoys higher correlation with FID: 0.97, while the correlation of CKA and FID is lower: 0.79.

\section{Details on experiments with convolutional networks}

\begin{table}[tbh]
\begin{tabular}{llllll}
\multicolumn{2}{l}{Metrics to correlate}                                   & Noise                           & Gaussian Blur                      & Grayscale                         & Hue                       \\ \hline
Disagreement & RTD     & \textbf{0.966 $\pm$ 0.001} & \textbf{0.982 $\pm$ 0.004} & \textbf{0.990 $\pm$ 0.004 } & \textbf{0.978 $\pm$ 0.008} \\
                             & $1-$CKA & 0.927 $\pm$ 0.006                            & 0.913 $\pm$ 0.011                           & 0.928 $\pm$ 0.040                            & 0.927 $\pm$ 0.017                           \\ \hline
Error rate   & RTD     & \textbf{0.982 $\pm$ 0.002}                           & 0.963 $\pm$ 0.007                            & 0.856 $\pm$ 0.052                           & 0.935 $\pm$ 0.030                           \\
                             & $1-$CKA & 0.966 $\pm$ 0.007 & \textbf{0.999 $\pm$ 0.001 } & \textbf{0.958 $\pm$ 0.018}  & \textbf{0.944 $\pm$ 0.033}
\end{tabular}
\label{tbl:shift_appendix_1}
 \caption{Analysis of ResNet-20 representations under different data distribution shifts. The correlation of RTD and $1-$CKA with Disagreement and Error rate.}
\end{table}

\begin{figure}[tbh]
\centering
\begin{subfigure}{.24\textwidth}
  \centering
  \includegraphics[width=1\linewidth]{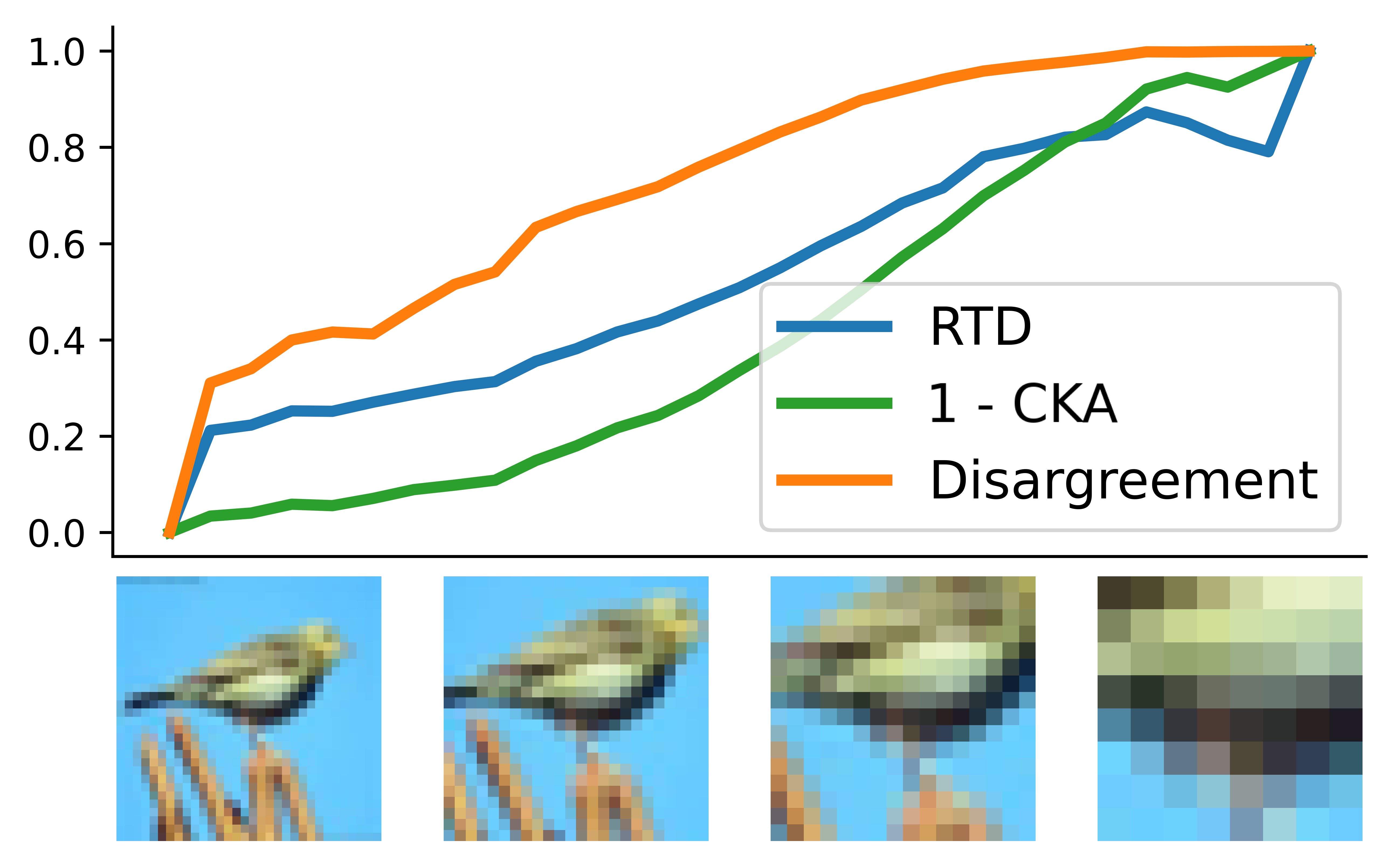}
  \caption{Zoom}
  \label{fig:zoom}
\end{subfigure}%
\begin{subfigure}{.24\textwidth}
  \centering
  \includegraphics[width=1\linewidth]{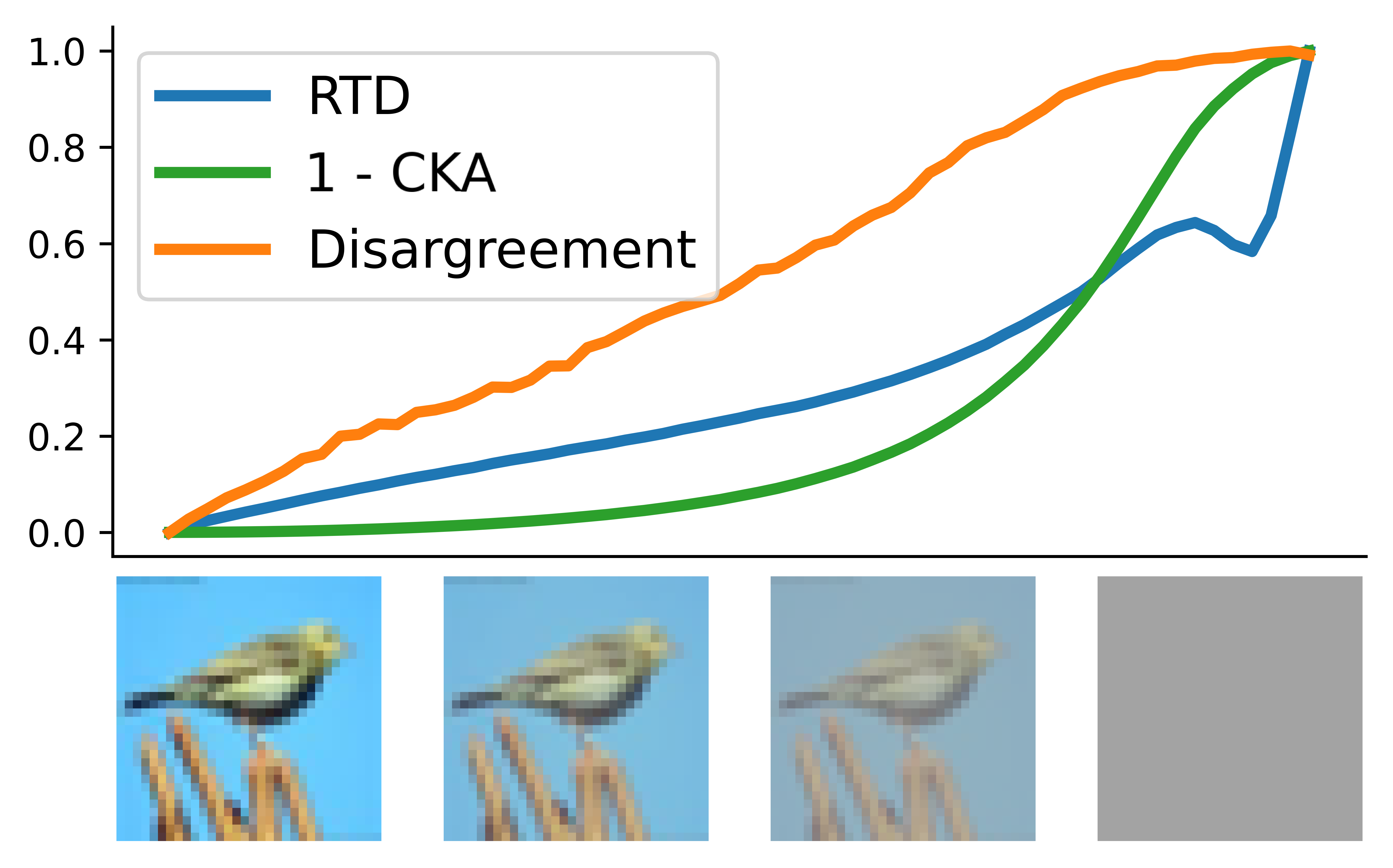}
  \caption{Contrast}
  \label{fig:contrast}
\end{subfigure}
\begin{subfigure}{.24\textwidth}
  \centering
  \includegraphics[width=1\linewidth]{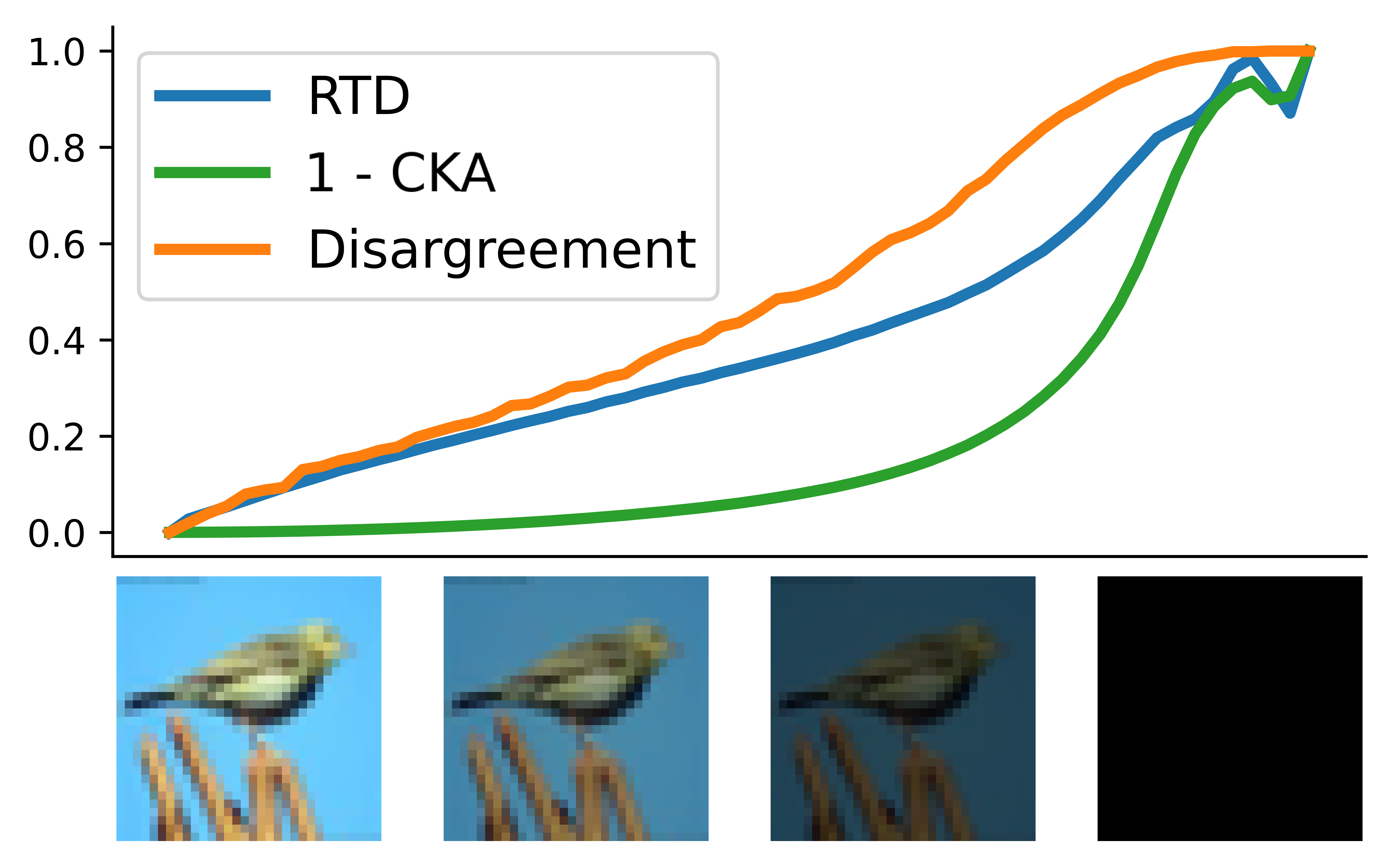}
  \caption{Brightness}
  \label{fig:brightness}
\end{subfigure}
\begin{subfigure}{.24\textwidth}
  \centering
  \includegraphics[width=1\linewidth]{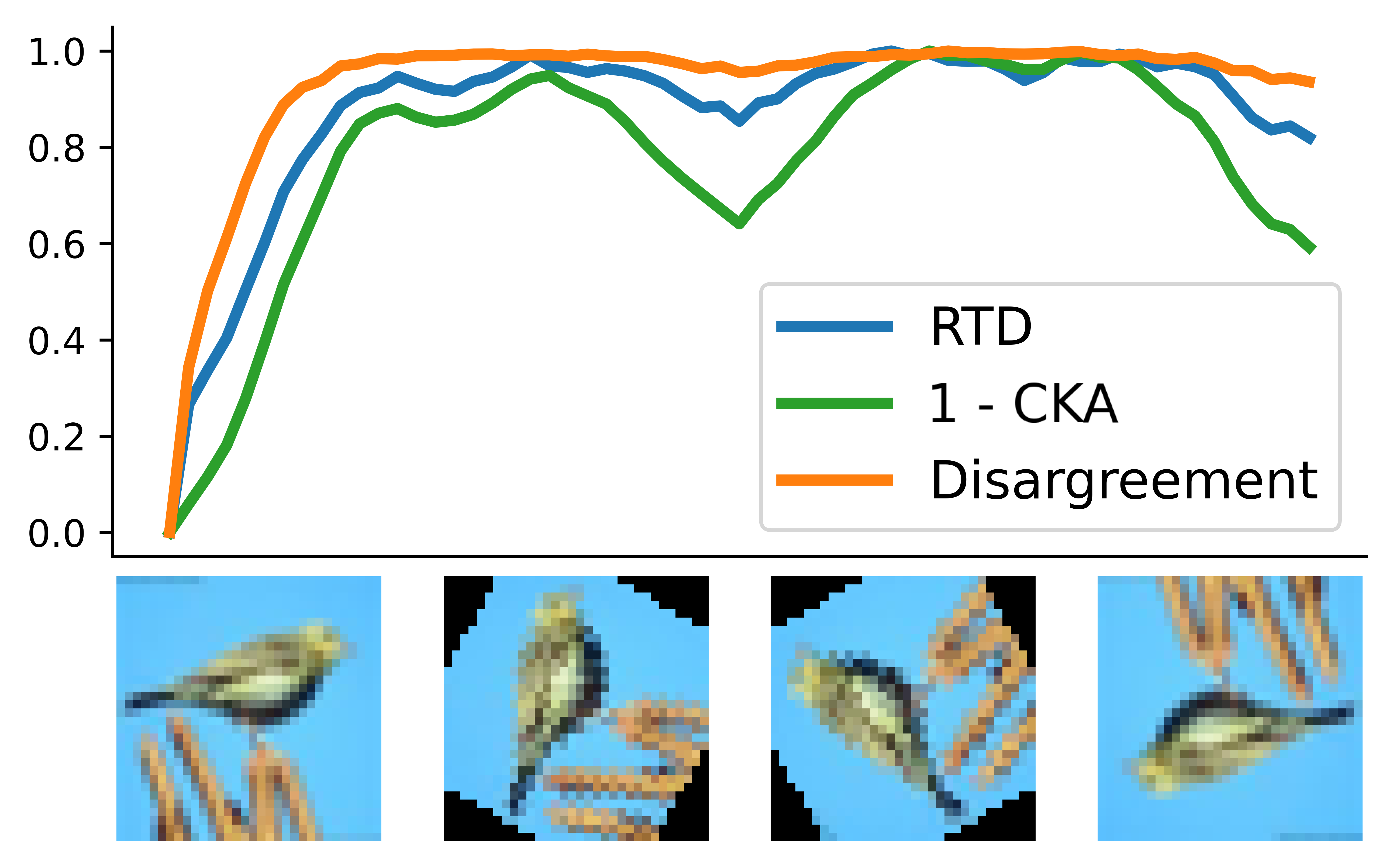}
  \caption{Rotation}
  \label{fig:rotation}
\end{subfigure}
\caption{Analysis of ResNet-20 representations under different data distribution shifts. The dynamics of scaled metrics with the monotonic application of various types of image transformations.}
\label{fig:shift_appendix}
\end{figure}

\begin{table}[tbh]
\begin{tabular}{llllll}
\multicolumn{2}{l}{Metrics to correlate}                                   & Zoom                            & Brightness                      & Contrast                         & Rotation                        \\ \hline
Disagreement & RTD     & \textbf{0.950 $\pm$ 0.006} & \textbf{0.975 $\pm$ 0.002} & \textbf{0.936 $\pm$ 0.010 } & \textbf{0.955 $\pm$ 0.015} \\
                             & $1-$CKA & 0.886 $\pm$ 0.010                           & 0.854 $\pm$ 0.024                           & 0.851 $\pm$ 0.021                            & 0.857 $\pm$ 0.020                           \\ \hline
Error rate   & RTD     & 0.946 $\pm$ 0.006                           & 0.921 $\pm$ 0.011                           & 0.937 $\pm$ 0.005                            & 0.940 $\pm$ 0.009                           \\
                             & $1-$CKA & \textbf{0.994 $\pm$ 0.002} & \textbf{0.997 $\pm$ 0.001} & \textbf{0.998 $\pm$ 0.001}  & \textbf{0.981 $\pm$ 0.005}
\end{tabular}
\label{tbl:shift_appendix_2}
 \caption{Analysis of ResNet-20 representations under different data distribution shifts. The correlation of RTD and $1-$CKA with Disagreement and Error rate.}
\end{table}

\begin{table}[tbh]
    \begin{subtable}[]{0.45\textwidth}
        \centering
        \begin{tabular}{lll}
                     & RTD & $1-$CKA \\ \hline
        Disagreement & \textbf{0.98 $\pm$ 0.01} & 0.93 $\pm$ 0.02     \\ 
        Error rate   & 0.9 $\pm$ 0.03& \textbf{0.99 $\pm$ 0.01}     \\ 
        \end{tabular}
       \caption{CIFAR-100}
       \label{tab:transfer_cifar100}
    \end{subtable}
    \hfill
    \begin{subtable}[]{0.45\textwidth}
        \centering
        \begin{tabular}{lll}
                     & RTD & $1-$CKA \\ \hline
        Disagreement & \textbf{0.91 $\pm$ 0.01}  & 0.89 $\pm$ 0.02    \\ 
        Error rate   & 0.60 $\pm$ 0.02& \textbf{0.73 $\pm$ 0.01}     \\ 
        \end{tabular}
        \caption{CIFAR-10}
        \label{tab:transfer_cifar10}
     \end{subtable}
     \caption{The correlation of metric dynamics when transferring the ResNet-20 network from CIFAR-100 to CIFAR-10 dataset.}
     \label{tab:transfer_appendix}
\end{table}

\begin{table}[tbh]
\centering
\begin{tabular}{lll}
                        & VGG-11                                                                                                                                 & \multicolumn{1}{r}{ResNet-20}                                                                                                                            \\ \hline
Number of epochs        & \multicolumn{2}{c}{100}                                                                                                                                                                                                                                                          \\ \hline
Optimizer               & \multicolumn{2}{c}{SGD, momentum=0.9}                                                                                                                                                                                                                                            \\ \hline
Learning rate (initial) & \multicolumn{2}{c}{0.1}                                                                                                                                                                                                                                                          \\ \hline
Scheduler               & \multicolumn{2}{c}{\begin{tabular}[c]{@{}c@{}}\textless 50\%: 0.1\\ 50-90\%: 0.1-0.001 (linear)\\ \textgreater 90\%: 0.001\end{tabular}} \\ \hline
Batch size              & \multicolumn{2}{c}{128}                                                                                                                                                                                                                                                          \\ 
\end{tabular}
\caption{Details on learning the neural networks from random initialization on CIFAR datasets.}
\label{fig:cifarsetup}
\end{table}

\begin{table}[tbh]
\centering
\begin{tabular}{lll}

                        & Encoder part & \multicolumn{1}{r}{Classifier part}                                                                                                            \\ \hline
Number of epochs        & \multicolumn{2}{c}{50}                                                                                                                 \\ \hline
Optimizer               & \multicolumn{2}{c}{SGD, momentum=0.9}                                                                   \\ \hline
Learning rate (initial) & 0.001        & \multicolumn{1}{r}{0.1}                                                                                                                        \\ \hline
Scheduler               & None         & \begin{tabular}[c]{@{}r@{}}\textless 50\%: 0.1\\ 50-90\%: 0.1-0.001 (linear)\\ \textgreater 90\%: 0.001\end{tabular} \\ \hline
Batch size              & \multicolumn{2}{c}{128}                                                                                                                  \\ 
\end{tabular}
\caption{Details on fine-tuning the ResNet-20 from CIFAR-100 to CIFAR-10 dataset.}
\label{fig:finetunesetup}
\end{table}

\section{Experiments with disentanglement}
\label{sec:disent}

\begin{table}
    \centering
    \caption{Evaluation of the disentanglement for various directions in the latent space of \texttt{dSprites}.}
    \begin{tabular}{cc}
    \textbf{axis} & \textbf{RTD}\\
    \hline
    axis 1 & 148.1 \\ 
    axis 2 & 71.3 \\
    axis 3 & 53.4 \\
    axis 4 & 41.2 \\
    axis 5 & 40.5\\
    random & $162.8 \pm 18.6$
    \end{tabular}
    \label{tbl:disent}
\end{table}

Learning disentangled representations is a fundamental problem for improving the generalization, robustness, and interpretability of generative models. 
Let $Z$ be a latent space, $X$ - a space of objects, $g: Z \to X$ - a generator. 
The disentanglement of generative models can be evaluated by comparing the topology of data manifold slices
  $X_v = g(Z\mid_{z_i=v})$ for different values of $v$ \citep{zhou2021evaluating,barannikov2022homological}. If the direction $z_i$ corresponds to an interpretable factor, then $X_v$ must be topologically similar for different $v$.

We use the following experimental design.
$Z_{v,n} = \{z \in Z \mid (z,n)=v\}$ - a slice in a latent space orthogonal to a unit vector $n$.
We take a finite random sample $Z_1 \subset Z_{v,n}$
and a shifted sample $Z_2 = \{z_i + \delta n\}_{i=1}^{|Z_1|}$ for small $\delta$.
By definition, $Z_1$ and $Z_2$ have natural point-wise mapping and we can estimate homological similarity of $g(Z_1)$ and $g(Z_2)$ by RTD. 

In this experiment, we use \texttt{dSprites}\footnote{\url{https://github.com/deepmind/dsprites-dataset}} for the evaluation of disentanglement. \texttt{dSprites} is a dataset of procedurally generated  2D shapes from 5 ground truth independent latent factors: shape, scale, rotation, x-position, and y-position of a sprite. Thus, the latent space is disentangled and fully factorized. 
Particularly, we compare the slices orthogonal to axis-aligned vectors and orthogonal to random vectors, see Table \ref{tbl:disent}.
Except for the first axis, the topological dissimilarity estimated by RTD is significantly less than for a random direction. The first axis corresponds to a categorical factor - shape for which the aforementioned approach is arguably not applicable.
The \texttt{dSprites} dataset is quite simple and RTD was calculated for point clouds in the pixel space. However, the same technique can be straightforwardly applied to evaluate the disentanglement of image representations for more complex datasets. 

\begin{figure}[t]
    \centering
    \includegraphics[width=0.5\textwidth]{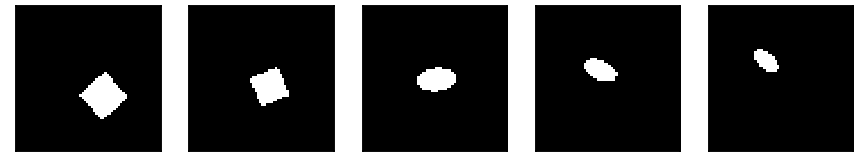}\\
    \includegraphics[width=0.5\textwidth]{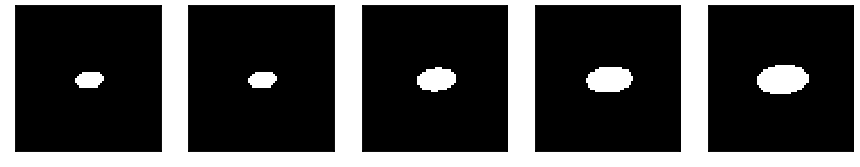}
    \caption{dSprites generated across directions in the latent space, top: random direction, bottom: axis-aligned direction, corresponds to an interpretable factor of variation.}
    \label{fig:dsprites}
\end{figure}

\section{Details on dimensionality reduction of MNIST with UMAP}
\label{app:mnist_umap}

\begin{figure}[th]
\centering
\includegraphics[width=0.19\textwidth]{img/umap_mnist1.png}
\includegraphics[width=0.19\textwidth]{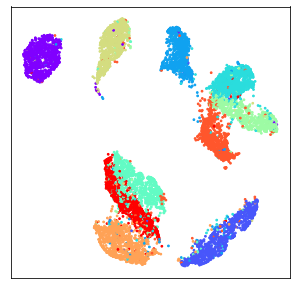}
\includegraphics[width=0.19\textwidth]{img/umap_mnist3.png}
\includegraphics[width=0.19\textwidth]{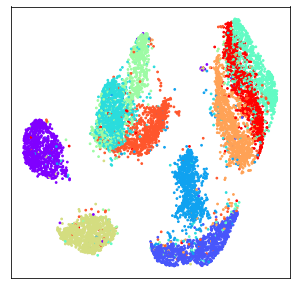}
\includegraphics[width=0.19\textwidth]{img/umap_mnist5.png}
\caption{2D representations of MNIST produced by UMAP, n\_neighbors $\in (10, 20, 50, 100, 200) $}
\label{fig:umap_mnist}
\end{figure}

Visual inspection of Figure \ref{fig:umap_mnist} reveals apparent incoherences of CKA. Denote by $U(n)$ representations obtained by UMAP with the number of neighbors $n$. According to CKA (Figure \ref{fig:umap_mnist_cka}), $U(10)$ is closer to $U(200)$ than to $U(20)$;
also $U(200)$ is closer to $U(10)$ than to $U(100)$.

\section{Additional experiments}
\label{app:additional}

For the ``clusters'' experiments, we did additional comparisons of point clouds with alternative similarity measures. Firstly, we calculated CKA with the RBF kernel for 3 bandwidths equal to 0.2, 0.4, 0.8 of median pairwise distances (as proposed in \cite{kornblith2019similarity}). The performance as measured by Kendall-tau correlation with the true ordering was 0.23, 0.04, 0.14 - not better than for CKA with the linear kernel. Secondly,
we applied the topological loss term from \cite{moor2020topological}. The performance as measured by Kendall-tau correlation with the true ordering was poor: -0.52.

\section{Internal similarity of Neural Network layers}
\label{app:imagenet_layers}
\begin{figure}[hb!]
    \centering
    \vskip-0.19in
    \includegraphics[width=0.9\textwidth]{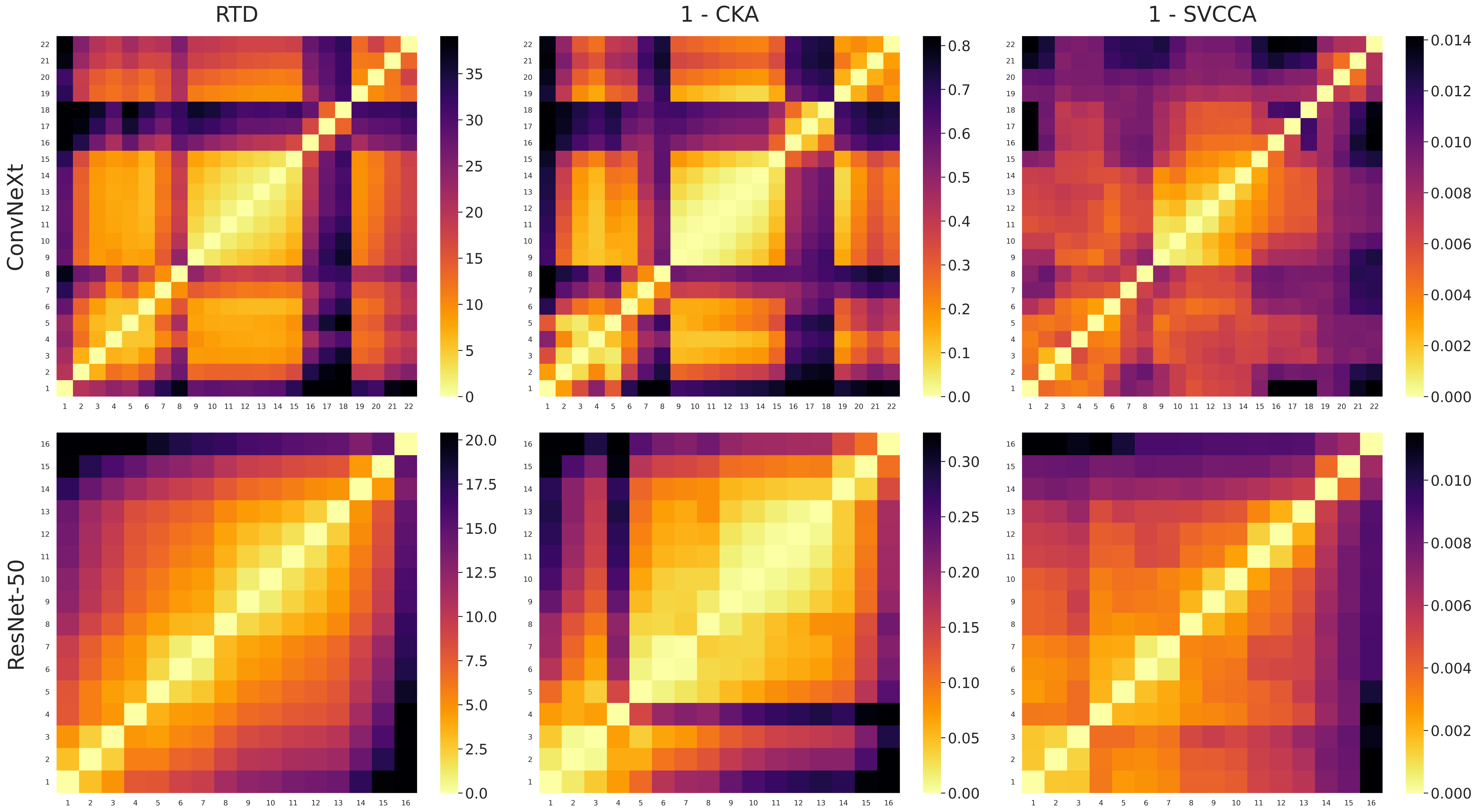}
    \caption{The representation differences between the layer blocks within trained networks,  ImageNet-1k dataset. The columns correspond to the metrics, and the rows -- to the architectures.}
    \label{fig:imagenet-layers}
\end{figure}

Here we compare the outputs of layer blocks within the trained network. We consider ResNet-50 and ConvNeXt-tiny \cite{liu2022convnet} architectures pre-trained on ImageNet-1k dataset \cite{deng2009imagenet}. We calculate RTD, CKA and SVCCA within outputs after each Bottleneck Residual Block or ConvNeXt's block respectively.
In Fig. \ref{fig:imagenet-layers}, we plot similarity between layers within each architecture. We observe that RTD catches architecture's block structure better than CKA, SVCCA. The ResNet-50 architecture has sequence of blocks in form [3, 4, 6, 3] and it can be seen that RTD highlights it with sub-squares of corresponding sizes.

\end{document}